\documentclass[10pt,twocolumn,letterpaper]{article}

\usepackage{iccv}
\usepackage{times}
\usepackage{epsfig}
\usepackage{graphicx}
\usepackage{amsmath}
\usepackage{amssymb}
\usepackage{booktabs}
\usepackage{bm}
\usepackage{pifont}

\usepackage[pagebackref=true,breaklinks=true,letterpaper=true,colorlinks,bookmarks=false]{hyperref}

\iccvfinalcopy

\usepackage[capitalize]{cleveref}
\crefname{section}{Sec.}{Secs.}
\Crefname{section}{Section}{Sections}
\Crefname{table}{Table}{Tables}
\crefname{table}{Tab.}{Tabs.}

\usepackage{marvosym}
\usepackage{balance}
\usepackage{arydshln}
\usepackage{multirow}
\usepackage{makecell}

\newcommand{\tablestyle}[2]{\setlength{\tabcolsep}{#1}\renewcommand{\arraystretch}{#2}\centering\footnotesize}
\newlength\savewidth\newcommand\shline{\noalign{\global\savewidth\arrayrulewidth
\global\arrayrulewidth 1pt}\hline\noalign{\global\arrayrulewidth\savewidth}}
\usepackage[table]{xcolor}

\ificcvfinal\fi

\newcommand*{\affaddr}[1]{#1}
\newcommand*{\affmark}[1][*]{\textsuperscript{#1}}

\makeatletter
\newcommand{\printfnsymbol}[1]{\textsuperscript{\@fnsymbol{#1}}}
\makeatother
\newcommand\scalemath[2]{\scalebox{#1}{\mbox{\ensuremath{\displaystyle #2}}}}

\DeclareMathOperator{\SE}{SE}
\DeclareMathOperator{\Sim}{Sim}
\DeclareMathOperator{\SO}{SO}

\definecolor{Gray}{gray}{0.5}

\definecolor{Highlight}{HTML}{39b54a}
\definecolor{Modify}{HTML}{2240F0}

\definecolor{Highlight}{HTML}{3DA600}

\newcommand{\cmark}{\color{darkgray}\ding{51}}
\newcommand{\xmark}{\color{darkgray}\ding{55}}

\begin{document}

\makeatletter
\def\thanks#1{\protected@xdef\@thanks{\@thanks
        \protect\footnotetext{#1}}}
\makeatother

\title{Root Pose Decomposition Towards Generic Non-rigid 3D Reconstruction\\with Monocular Videos}

\author{
	Yikai Wang\affmark[1]\quad Yinpeng Dong\affmark[1,2]\quad\vspace{0.05in}Fuchun Sun\affmark[1]\textsuperscript{\Letter}$\thanks{\textsuperscript{\Letter}~Corresponding author: Fuchun Sun.}$\quad Xiao Yang\affmark[1]\\
	\affaddr{\affmark[1]Beijing National Research Center for Information Science and Technology$\,$(BNRist),\\ State Key Lab on Intelligent Technology and Systems,\\ Department of Computer Science and Technology, Tsinghua University}\quad
	\affaddr{\affmark[2]RealAI}\\
	\tt\small{\{yikaiw,dongyinpeng,fcsun\}@tsinghua.edu.cn, yangxiao19@mails.tsinghua.edu.cn}\\
}

\maketitle

\ificcvfinal\thispagestyle{empty}\fi

\begin{abstract}
This work focuses on the 3D reconstruction of non-rigid objects based on monocular RGB video sequences.  Concretely, we aim at building high-fidelity models for generic object categories and casually captured scenes. To this end, we do not assume known root poses of objects, and do not utilize category-specific templates or dense pose priors. The key idea of our method, Root Pose Decomposition (RPD),  is to maintain a per-frame root pose transformation,  meanwhile building a dense field with local transformations to rectify the root pose. The optimization of local transformations is performed by point registration to the canonical space. We also adapt RPD to multi-object scenarios with object occlusions and individual differences.  As a result, RPD allows non-rigid 3D reconstruction for complicated scenarios containing objects with large deformations, complex motion patterns, occlusions, and scale diversities of different individuals. Such a pipeline potentially scales to diverse sets of objects in the wild. We experimentally show that RPD surpasses state-of-the-art methods on the challenging DAVIS, OVIS, and AMA datasets.  We provide  video results in \url{https://rpd-share.github.io}.
\vskip-0.01in
\end{abstract}

\section{Introduction}
\label{sec:intro}

The reconstruction of non-rigid (or deformable) 3D objects with monocular RGB videos is a long-standing and challenging task in computer vision and graphics~\cite{DBLP:journals/corr/abs-2210-15664}. It is needed for a variety of applications ranging from XR to robotics.  Traditionally, a typical prior pipeline leverages template-based models such as human skeleton models SMPL~\cite{DBLP:journals/tog/LoperM0PB15}, SMPL-X~\cite{DBLP:conf/cvpr/PavlakosCGBOTB19}, GHUM(L)~\cite{DBLP:journals/tog/YangSZZ19}, and reconstruct models for specific categories like the human body or human face. These methods do not scale to diverse categories. With the success of Neural Radiance  Field (NeRF)~\cite{DBLP:conf/eccv/MildenhallSTBRN20}, several representative works~\cite{DBLP:journals/corr/abs-2206-15258,DBLP:conf/iccv/ParkSBBGSM21} unify  frames to a  canonical space and do not rely on pre-defined skeleton models.  Whilst a variety of methods have been proposed to improve the reconstruction fidelity, the performance degenerates when encountering large object deformations or movements. Meanwhile, they are not suitable for casual videos when background Structure from Motion (SfM) does not provide root poses for the object.  As a typical work to address the issues, BANMo~\cite{DBLP:conf/cvpr/YangVNRVJ22} initializes approximate camera poses by leveraging continuous surface embeddings (CSE)~\cite{DBLP:conf/nips/NeverovaNSKLV20} or pre-trained DensePose models~\cite{DBLP:conf/cvpr/GulerNK18}. Nevertheless, CSE is acquired by annotations and only applies to specific categories in the training set, \emph{e.g.}, quadruped animals. DensePose is learned with the aid of manual UV fields from SMPL and is thus also highly category-limited. Such a pipeline that relies on off-the-shelf pose or surface models of a certain categories  does not generalize to reconstruct generic object categories.  

As a result, it is an open problem when considering the non-rigid construction in the wild that might contain complicated factors, \emph{e.g.}, multiple categories, complex motion patterns, individual diversities, or  object occlusions. Based on casually captured monocular videos, our effort is devoted to building articulated models for generic categories, without explicitly incorporating priors that might limit the generalization of categories. Towards this goal, we  propose \textbf{Root Pose Decomposition (RPD)}, a method for  non-rigid 3D reconstruction based on monocular RGB videos. RPD does not rely on known camera poses or poses compensated by background-SfM, category-specific skeletons, or  pre-trained dense pose models (\emph{e.g.}, DensePose, CSE), while could achieve articulated  reconstruction for objects with rapid object deformations, complex motion patterns, and large pose changes.

\begin{table*}[h]
	\centering
 \vskip -0.06in
	\tablestyle{0.5pt}{0.7}
	\resizebox{0.8\linewidth}{!}{
		\begin{tabular}{lcccc}
			\toprule
	Method & \textbf{Nerfies}~\cite{DBLP:conf/iccv/ParkSBBGSM21} & \textbf{ViSER}~\cite{DBLP:conf/nips/YangSJVCLR21}& \textbf{BANMo}~\cite{DBLP:conf/cvpr/YangVNRVJ22} & \textbf{Ours}\\
        \midrule
        \textbf{Dependency (\xmark$\;$is preferred)}&&&&\\        
        Known camera poses&\xmark &\xmark&\xmark&\xmark\\
        Background-SfM$^{*}$ &\cmark &\xmark&\xmark&\xmark\\
        Pre-trained CSE/DensePose$^{{\ddagger}}$ &\xmark&\xmark&\cmark&\xmark\\
        \midrule
        \textbf{Algorithm}&&&&\\
        Canonical space sharing$^{*\dagger{\ddagger}}$ & Multi-frame&Multi-frame&Multi-frame&Multi-frame \& multi-object\\
        Warping function& \cmark&\cmark&\cmark&\cmark\\ 
        Dense transformation field$^{*\dagger{\ddagger}}$  & $\SE(3)$ &\xmark & \xmark &$\Sim(3)$ \\
        Linear skinning weights$^{*}$&\xmark&\cmark&\cmark&\cmark\\
        Registration$^{*\dagger{\ddagger}}$&Photometric$\;$&Self-supervised feature&CSE feature&Non-rigid point registration\\
        Shape representation$^{\dagger}$ & Implicit & Mesh & Implicit & Implicit\\
        \midrule
        \textbf{Performance (\cmark$\;$is preferred)} &&&&\\
        Handling large deformation$^{*\dagger}$ &\xmark &\xmark&\cmark&\cmark\\    
        Reconstruction for generic  categories $^{{\ddagger}}$ &\cmark &\cmark&\xmark&\cmark\\        
        Multi-object  occlusions \& differences$^{*\dagger{\ddagger}}$ &\xmark &\xmark&\xmark&\cmark\\      	
        \shline
		\end{tabular}
	}
\vskip0.04in
\caption{Comparison with related methods. We highlight the differences with Nerfies, ViSER, and BANMo by $^{*}$, $^{\dagger}$, and {$^{\ddagger}$}, respectively.}
\label{table:compare}
\vskip-0.03in
\end{table*}

Concretely, RPD follows the common approach for non-rigid reconstruction that builds a canonical space for different frames with learned warping functions. Here, the warping function consists of a mapping function and also a shared root pose transformation per frame. The core idea of RPD lies in its decomposition of the root pose into local transformations consisting of rotations, translations, and scaling factors at observation points by building a dense $\Sim(3)$ field. On the one hand, unlike the rigid root pose,  incorporating the local transformation provides more room for each point to fit the canonical model but simultaneously keeps the local transformation continuous with position, direction, and time. On the other hand, maintaining a shared root pose per frame stabilizes the learning process of warping functions. 

Besides, towards the generic reconstruction in the wild, we propose techniques  that are naturally compatible with RPD. As a result, RPD could be applied to multi-object scenarios with more challenging occluded objects. Apart from that, RPD also adapts to the complex scenes with different individuals that vary in shape, height, or scale. By category-level generalization and the ability to adapt to individual differences, we could reconstruct multiple objects simultaneously and no longer need to collect multiple video sequences that contain the exactly same object individual.

Experiments for evaluating RPD are conducted on three monocular video datasets, including two challenging multi-object segmentation datasets DAVIS~\cite{DBLP:journals/corr/abs-1905-00737} and OVIS~\cite{DBLP:journals/ijcv/QiGHWLBBYTB22},  a reconstruction dataset AMA~\cite{DBLP:journals/tog/VlasicBMP08} with ground truth meshes for evaluation, and a casually collected dataset Casual~\cite{DBLP:conf/cvpr/YangVNRVJ22}. Experimental results demonstrate the superiority of our design and   the potential towards reconstructing large-scale generic object categories with  monocular RGB videos.

A  comparison with related methods is added in Table~\ref{table:compare}. To summarize, the contributions of our work are:
\begin{itemize}
\item We reveal that estimating per-frame root poses is a core factor for  the non-rigid 3D reconstruction of generic categories, with few dependencies on category-specific templates/pose priors, known camera poses, or poses compensated by background-SfM. 

\item With this motivation, we propose an effective method (RPD) to estimate the per-frame root pose by decomposing  the root pose into local transformations, and for the first time, we propose to leverage the success of the non-rigid point registration field for the non-rigid  monocular 3D reconstruction. 

\item Apart from the promising results achieved by RPD, the method is  compatible with handling multiple objects  with occlusions and individual differences.
\end{itemize}

\section{Related Work}
\label{sec:related_work}

We introduce recent methods of monocular non-rigid 3D reconstruction and non-rigid point cloud registration, since they are both related to our paper. 

\textbf{Monocular non-rigid reconstruction.} A large number of works focus on this topic. For example, shape from template (SfT)~\cite{DBLP:journals/pami/BartoliGCCP15,DBLP:journals/ijcv/GallardoPCB20,DBLP:conf/cvpr/KairandaTETG22} assumes a static template is given as a prior. Non-rigid structure from motion (NRSfM)~\cite{DBLP:conf/cvpr/BreglerHB00,DBLP:conf/iccv/KongL19,DBLP:conf/3dim/RussellFA12} constructs objects without using 3D template priors, but might heavily rely on observed point trajectories~\cite{DBLP:journals/ijcv/SandT08,DBLP:conf/eccv/SundaramBK10}. We refer readers to the survey~\cite{DBLP:journals/corr/abs-2210-15664} for a more detailed overview.
Neural rendering methods are more related to ours. To build a higher-fidelity canonical space based on NeRF~\cite{DBLP:conf/eccv/MildenhallSTBRN20}, Nerfies~\cite{DBLP:conf/iccv/ParkSBBGSM21} formulates warping functions into a dense $\SE(3)$ field that encodes rigid motions with elastic regularization to encourage rigid local deformations. NDR~\cite{DBLP:journals/corr/abs-2206-15258} designs bijective mappings with a strictly invertible representation optimized given mainly RGB-D inputs. Nevertheless,  many typical approaches ~\cite{DBLP:journals/corr/abs-2206-15258,DBLP:conf/cvpr/LiNSW21,DBLP:conf/iccv/ParkSBBGSM21,DBLP:journals/tog/ParkSHBBGMS21} assume that object movements are small and camera transformations are given, since learning warping functions is especially hard when encountering large pose changes and object movements, such as running. BANMo~\cite{DBLP:conf/cvpr/YangVNRVJ22} models warping functions with linear skinning weights and root poses obtained by off-the-shelf templates (\emph{e.g.}, SMPL~\cite{DBLP:journals/tog/LoperM0PB15}). However, the current pipeline with  root poses highly relies on
category-specific priors like CSE surface features~\cite{DBLP:conf/nips/NeverovaNSKLV20} (\emph{e.g.}, for cats). Different from existing methods, our  method  handles large object deformations and movements without needing known camera poses, background-SfM, or pre-trained dense pose/CSE  models.

\textbf{Root pose estimation.} For objects in monocular videos, ``root pose'' presents the global 3D orientation of each  target object. It remains tough to estimate  root poses given the ambiguity in reflective symmetry, textureless region, etc. ShSMesh~\cite{DBLP:conf/cvpr/YeT021} adopts view synthesis adversarial learning to optimize camera poses. U-CMR~\cite{DBLP:conf/eccv/GoelKM20} proposes camera-multiplex that represents the distribution over cameras, which is yet category-specific. Both methods are designed for rigid objects. For non-rigid objects, NRSfM-related works~\cite{DBLP:conf/cvpr/BreglerHB00,DBLP:conf/cvpr/DaiLH12,DBLP:journals/ijcv/TomasiK92} estimate poses from 2D point trajectories in a class-agnostic manner, but is weak at capturing long-range correspondences or estimating root poses in the wild. DOVE~\cite{DBLP:journals/corr/abs-2107-10844} adopts view space shape reflection. BANMo~\cite{DBLP:conf/cvpr/YangVNRVJ22} relies on pre-trained CSE~\cite{DBLP:conf/nips/NeverovaNSKLV20} and DensePose~\cite{DBLP:conf/cvpr/GulerNK18} to initialize root poses, and  the category is limited to human or quadruped animals. Hence, BANMo does not apply to generic object categories and is prone to be affected by the transferred quality of pre-built surface/pose embeddings from large-scale datasets. A contemporary work L2G-NeRF~\cite{DBLP:journals/corr/abs-2211-11505} predicts  accurate camera pose for  bundle-adjusting NeRFs. ViSER~\cite{DBLP:conf/nips/YangSJVCLR21}  adopts optical flow that learns approximate pixel-surface embeddings for pose initialization. We have a similar pose initialization scheme as ViSER. Differently, we rectify  root poses during training by non-rigid point registration.

\textbf{Non-rigid point registration.} This task aims to build a deformation field or estimate the point-to-point alignment  from one point cloud to another~\cite{DBLP:journals/cgf/DengYDZ22}. For example, deformation graph~\cite{DBLP:journals/tog/SumnerSP07} builds a sparsely sub-sampled graph from the surface and propagates deformation from node to surface. Lepard~\cite{DBLP:conf/cvpr/LiH22}  learns partial point cloud mapping for guiding the global non-rigid registration~\cite{DBLP:conf/nips/VaswaniSPUJGKP17}. DeformationPyramid~\cite{DBLP:journals/corr/abs-2205-12796}  creates a multi-level  deformation motion field for registration in general scenes. This work leverages the idea of non-rigid point registration to learn the decomposed transformation at each observation point.

\section{Methodology}
\label{sec:methodology}

In this section, we first introduce the basic framework for non-rigid 3D reconstruction with monocular RGB videos in Sec.~\ref{subsec:deform_frame}. We describe our proposed  RPD, a method for handling large movements and diverse poses of non-rigid objects  in Sec.~\ref{subsec:quadratic_pose_field}, followed by its optimization and  registration strategies in Sec.~\ref{subsec:optimization}.

\subsection{Preliminary: Non-rigid Reconstruction}
\label{subsec:deform_frame} 

Prior to going further, we first provide the notations
 and necessary preliminaries used in our paper. Given sequences of monocular RGB videos with  totally $T$ frames that consist of one or multiple object identities for reconstruction, different object identities of the same category (\emph{e.g.}, human)  in different frames are supposed to share a common canonical space. To achieve this, each 3D point $\mathbf{x}_*\in\mathbb{R}^3$ in the canonical space   corresponds to a 3D point $\mathbf{x}_t\in\mathbb{R}^3$  in the camera space from the  $t$-th frame (termed as time $t$) within the $T$ frames. Here, $\mathbf{x}_t$ locates on the ray which emanates from a 2D pixel in the frame image. Inspired by the recent non-rigid reconstruction methods~\cite{DBLP:journals/corr/abs-2206-15258,DBLP:conf/iccv/ParkSBBGSM21,DBLP:conf/cvpr/YangVNRVJ22}, we learn time-variant warping functions $\mathcal{W}_{t\to *}$ and $\mathcal{W}_{*\to t}$, satisfying
\begin{align}
\mathbf{x}_*=\mathcal{W}_{t\to *}(\mathbf{x}_t),\;\;\mathbf{x}_t=\mathcal{W}_{*\to t}(\mathbf{x}_*).
\label{eq:warping}
\end{align}

\begin{figure*}[t!]
\centering
\includegraphics[width=0.95\textwidth]{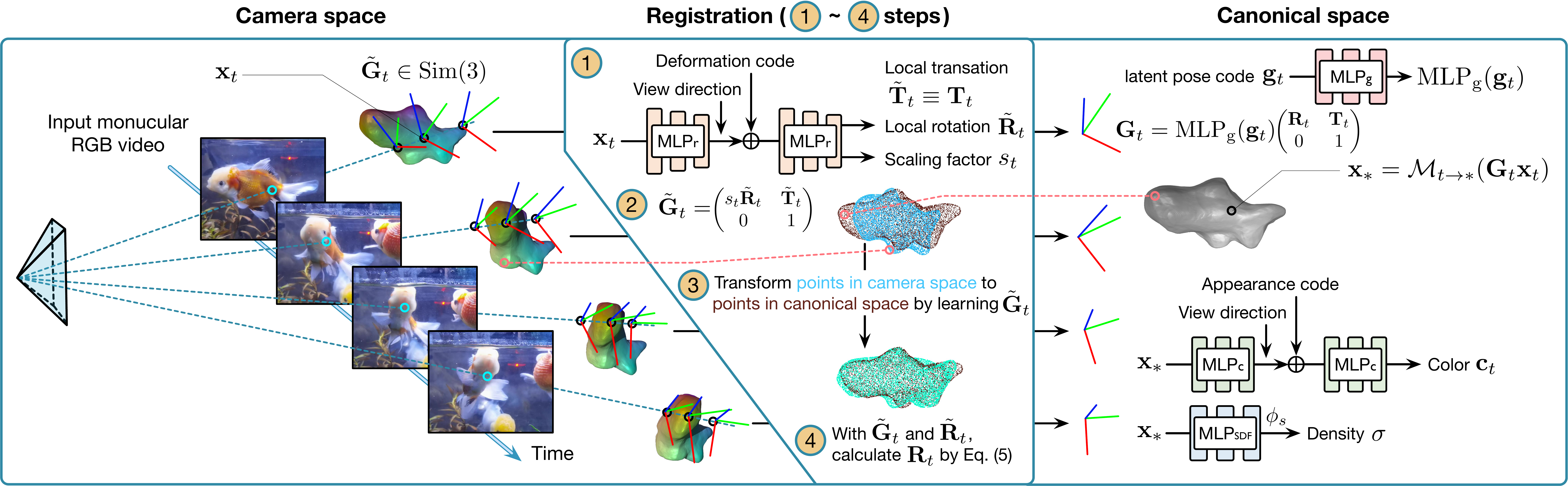}
\caption[]{Overall architecture of our RPD for non-rigid monocular reconstruction. In RPD, the root pose transformation  $\mathbf{G}_t$ is computed by local transformations $\tilde{\mathbf{G}}_t$ at each point $\mathbf{x}_t$. Decomposing $\mathbf{G}_t$ into a dense $\Sim(3)$ field allows more flexible registration to the canonical space. While maintaining a per-frame global $\mathbf{G}_t$ instead of directly using $\tilde{\mathbf{G}}_t$ helps the optimization of  $\mathcal{M}_{t\to *}$ more stable.}
\label{pic:method}
\vskip-0.07in
\end{figure*}

Predicting articulated 3D models for non-rigid objects is now translated to  reconstructing a  static  canonical space, where NeRF~\cite{DBLP:conf/eccv/MildenhallSTBRN20} could be applied to  predict  two properties of $\mathbf{x}_*$ with Multi-Layer Perception (MLP),  including its color $\mathbf{c}_t\in \mathbb{R}^3$ and  density $\sigma\in [0,1]$, 
\begin{align}
\mathbf{c}_t=\text{MLP}_\text{c}(\mathbf{x}_*,\mathbf{d}_t, \bm{\omega}_t), \;\; \sigma=\phi_s\big(\text{MLP}_\text{SDF}(\mathbf{x}_*)\big),
\label{eq:nerf}
\end{align}
where $\mathbf{d}_t\in\mathbb{R}^2$ and $\bm{\omega}_t$ denote  the viewing direction and the latent appearance  code~\cite{DBLP:conf/cvpr/Martin-BruallaR21} at time $t$, respectively. The object surface is extracted as the zero level-set of signed distance function (SDF), with $\phi_s(\cdot)$ being a unimodal density distribution to convert SDF  to the density~\cite{DBLP:conf/nips/WangLLTKW21}.

When given sampled $\mathbf{x}_t$ from the camera space,  the main optimization is performed based on $\mathcal{W}_{t\to *}$ and Eq.~(\ref{eq:nerf}) by minimizing the color reconstruction loss and the silhouette reconstruction loss, following existing volume rendering pipelines~\cite{DBLP:conf/eccv/MildenhallSTBRN20,DBLP:conf/nips/YarivKMGABL20}. Besides, with the help of $\mathcal{W}_{*\to t}$, a cycle loss could be introduced to maintain the cycle consistency between deformed frames~\cite{DBLP:journals/corr/abs-2206-15258,DBLP:conf/cvpr/LiNSW21}.

In summary, the reconstruction of non-rigid objects combines building the canonical space and NeRF. Yet the way to determine warping functions remains challenging and attracts much research effort. As described in Sec.~\ref{sec:related_work}, typical approaches include  formulating $\mathcal{W}_{t\to *}$ into a dense $\SE(3)$ field~\cite{DBLP:conf/iccv/ParkSBBGSM21}, or using a hyper-embedding if the surface undergoes topology changes~\cite{DBLP:journals/corr/abs-2206-15258,DBLP:journals/tog/ParkSHBBGMS21},  usually under the assumption of small pose changes, small object movements, and known (or  by background-SfM) camera parameters ~\cite{DBLP:journals/corr/abs-2206-15258,DBLP:conf/cvpr/LiNSW21,DBLP:conf/iccv/ParkSBBGSM21,DBLP:journals/tog/ParkSHBBGMS21}. Meanwhile, the reconstruction pipeline with pre-estimated root poses~\cite{DBLP:conf/cvpr/YangVNRVJ22}  requires reliable category-specific priors like DensePose CSE features~\cite{DBLP:conf/nips/NeverovaNSKLV20}.

\subsection{Root Pose Decomposition}
\label{subsec:quadratic_pose_field}

Supposing the warping function $\mathcal{W}_{t\to *}$ is composed by a mapping function $\mathcal{M}_{t\to *}$ and a root transformation matrix $\mathbf{G}_t\in\SE(3)$ that uniformly rotates and translates   the points at time $t$ to the canonical space. A similar decomposition is also applied to $\mathcal{W}_{*\to t}$. Specifically, by Eq.~(\ref{eq:warping}), there is, 
\begin{align}
\mathbf{x}_*=\mathcal{M}_{t\to *}(\mathbf{G}_t\mathbf{x}_t), \;\;\mathbf{x}_t=\mathbf{G}_t^{-1}\mathcal{M}_{*\to t}(\mathbf{x}_*).
\label{eq:warping2}
\end{align}

Optimizing a separate root transformation  $\mathbf{G}_t$ is to handle large  deformations and movements of non-rigid objects in monocular videos. Intuitively, at time $t$, we expect that $\mathbf{G}_t$ rotates and translates all observed points by fitting the corresponding points in the canonical space to a large extent, making it more focused when learning $\mathcal{M}_{t\to *}$. To estimate $\mathbf{G}_t$,  existing methods such as PnP solutions might suffer from catastrophic failures given the non-rigidity of objects, and CSE-based pose estimation methods~\cite{DBLP:conf/cvpr/YangVNRVJ22} limit the category-level generalization of the reconstruction. In view of this, we propose  Root Pose Decomposition (RPD) that learns the root transformation matrix without needing pre-built continuous surface features or camera transformations. An overall architecture of  RPD is depicted in Fig.~\ref{pic:method}.

We let the root transformation  $\mathbf{G}_t=\text{MLP}_\text{g}(\mathbf{g}_t)\scalemath{0.7}{
\begin{pmatrix}
\mathbf{R}_t&\mathbf{T}_t \\
0 & 1 
\end{pmatrix} }$ consist of a rotation  $\mathbf{R}_t\in\SO(3)$ and a translation  $\mathbf{T}_t\in\mathbb{R}^{3}$, where $\mathbf{g}_t$ denotes the latent pose code. Specifically, $\mathbf{g}_t$ is time-variant and represented by a Fourier embedding~\cite{DBLP:conf/eccv/MildenhallSTBRN20}. $\text{MLP}_\text{g}$   takes as input  the Fourier embedding and predicts poses with a rotation-translation head. At time $t$, we  parametrize $\mathbf{G}_t$ with the composition of individual local transformations $\tilde{\mathbf{G}}_t=\scalemath{0.7}{
\begin{pmatrix}
s_t\tilde{\mathbf{R}}_t&\tilde{\mathbf{T}}_t\\
0 & 1 
\end{pmatrix} }\in\Sim(3)$ at 3D points $\mathbf{x}_t$ in the camera space. Here,  $\tilde{\mathbf{R}}_t$ and $\tilde{\mathbf{T}}_t$ represent the local rotation and translation at $\mathbf{x}_t$, respectively. $s_t\in\mathbb{R}^+$ denotes a scaling factor that deals with scale variances of different individuals, as described in Sec.~\ref{subsec:step2general}. We  experimentally let $\tilde{\mathbf{T}}_t\equiv\mathbf{T}_t$ for all points but learn the per-point rotation matrix $\tilde{\mathbf{R}}_t\in\SO(3)$ and the scaling factor $s_t$.

Suppose $\mathbf{o}_*$ is the object center in the canonical space, which could be  approximately obtained  by averaging sampled points on the object surface given  SDF. The corresponding object center at time $t$ is then  computed by $\mathbf{o}_t=\mathcal{W}_{*\to t}(\mathbf{o}_*)$. We experimentally find that setting  $\mathbf{o}_t$ to a fixed point already achieves  promising  performance. We employ a NeRF-style  architecture that builds a dense $\Sim(3)$ field  to calculate the rotation matrix $\tilde{\mathbf{R}}_t$ of each point $\mathbf{x}_t$ around the center $\mathbf{o}_t$ and also estimate the scaling factor $s_t$, namely,
\begin{align}
\tilde{\mathbf{R}}_t,s_t=\text{MLP}_\text{r}(\mathbf{x}_t-\mathbf{o}_t,\mathbf{d}'_t, \bm{\varphi}_t),
\label{eq:rotate}
\end{align}
where $\mathbf{d}'_t\in\mathbb{R}^2$ is the viewing direction of the vector $\mathbf{x}_t-\mathbf{o}_t$ at time $t$, and $\bm{\varphi}_t$ is the  latent deformation code at time $t$.

Inspired by the hierarchical  architecture for point registration~\cite{DBLP:journals/corr/abs-2205-12796}, we parameterize $\text{MLP}_\text{r}$ in Eq.~(\ref{eq:rotate}) 
by a multi-level network where each level  outputs the motion increments from its previous level. More details of its optimization will be provided in Sec.~\ref{subsec:optimization}.

Denote  $\mathbf{x}_t^{n}$ as the $n$-th 3D point sampled along the camera ray that emanates from the 2D frame pixel, and denote $i$ as the  index of a ray. We  approximate $\mathbf{R}_t$ by considering the  rotation matrices  at  sampled points of all rays. Since multiple works~\cite{DBLP:journals/tog/BouazizMLKP14,DBLP:journals/tog/ChaoPSS10,DBLP:conf/iccv/ParkSBBGSM21} utilize the Frobenius norm to penalize the deviation from the closest rotation, inspired by this, we attain $\mathbf{R}_t$ by seeking a rotation matrix  that satisfies,
\begin{align}
\mathbf{R}_t=\mathop{\arg\min}\limits_{\mathbf{R}\in\SO(3)}
\sum_{i,n} \tau^n \big\|\tilde{\mathbf{R}}_t^n- \mathbf{R}\big\|_{\text{F}}^2,
\label{eq:mini_rotation}
\end{align}
where $\|\cdot\|_\text{F}$ denotes the Frobenius norm, and $\tau^n$ represents the probability of $\mathbf{x}_t^n$ being visible to the camera, calculated by $\tau^n=\prod_{m=1}^{n-1}\text{exp}(-\sigma^m\delta^m)(1-\text{exp}(-\sigma^n\delta^n))$ with the density $\sigma^n=\sigma(\mathcal{W}_{t\to *}(\mathbf{x}_t^n))$  by Eq.~(\ref{eq:nerf}) and  the interval $\delta^n$ between the $n$-th  sampled point and the next. Introducing  $\tau^n$ to Eq.~(\ref{eq:mini_rotation}) encourages a point near the surface to acquire greater importance. Here, the parameters  (\emph{e.g.}, $\tau^n$, $\tilde{\mathbf{R}}_t^n$) are also related with   $i$, and we omit the script for simplicity.

\subsection{Optimization and Registration}
\label{subsec:optimization}

Inspired by BANMo~\cite{DBLP:conf/cvpr/YangVNRVJ22} that reduces the reconstruction complexity, we apply linear skinning weights with control points to represent $\mathcal{M}_{t\to *}$ and $\mathcal{M}_{*\to t}$ in Eq.~(\ref{eq:warping2}). Here, we mainly focus on  optimizing  the root transformation and decomposed local transformations. 

Let $\mathbf{M}_t\in\mathbb{R}^{3\times 3}$ be the minimum point of  the summation part in Eq.~(\ref{eq:mini_rotation}) regardless of the $\SO(3)$ restriction. We first calculate $\mathbf{M}_t$ and then apply the singular-value decomposition on $\mathbf{M}_t$. To consider  $\SO(3)$, we compute $\mathbf{R}_t$ as the closest rotation matrix to $\mathbf{M}_t$ based on its decompositions,
\begin{gather}
\mathbf{M}_t=\frac{1}{\sum_{i,n}\tau^n}\sum_{i,n} \tau^n\tilde{\mathbf{R}}_t^n,\;\;\mathbf{M}_t=\mathbf{U}_t\mathbf{\Sigma}_t\mathbf{V}_t^\text{T},\\
\mathbf{R}_t=\mathop{\arg\min}\limits_{\mathbf{R}\in\SO(3)}\big\|\mathbf{M}_t-\mathbf{R}\big\|_\text{F}^2=\mathbf{U}_t\mathbf{V}_t^\text{T}.
\label{eq:singular_value}
\end{gather}

To learn the per-point rotation matrix $\tilde{\mathbf{R}}_t$ and the scaling factor $s_t$, we sample 
a set (denoted by $\mathbb{S}$) of points in the camera space, where the points with high $\tau^n$ values acquire large sampling probabilities. Inspired by point registration methods (see Sec.~\ref{sec:related_work} for more details), we encourage $\mathbb{S}$ to be  close to a set (denoted by $\mathbb{T}$) of  sampled surface points in the canonical space by leveraging the chamfer distance,
\begin{align}
\begin{split}
\mathcal{L}_\text{cd}&=\frac{1}{|\mathbb{S}|}\sum_{\mathbf{x}_t^n\in\mathbb{S}}\min_{\mathbf{x}_*\in\mathbb{T}}\Big\|\mathcal{M}_{t\to *}(\tilde{\mathbf{G}}_t^n\mathbf{x}_t^n)-\mathbf{x}_*\Big\|\\
&+\frac{1}{|\mathbb{T}|}\sum_{\mathbf{x}_*\in\mathbb{T}}\min_{\mathbf{x}_t^n\in\mathbb{S}}\Big\|\mathcal{M}_{t\to *}(\tilde{\mathbf{G}}_t^n\mathbf{x}_t^n)-\mathbf{x}_*\Big\|,
\label{eq:chamfer}
\end{split}
\end{align}
where the $l_1$ norm is adopted for the better partial-to-partial registration. By Eq.~(\ref{eq:chamfer}), we optimize $\tilde{\mathbf{R}}_t^n$ and $s_t$ while leaving $\tilde{\mathbf{T}}_t^n$ and $\mathcal{M}_{t\to *}$ temporally frozen.

Following DeformationPyramid~\cite{DBLP:journals/corr/abs-2205-12796}, we   use a deformability regularization  that encourages as-rigid-as-possible movement.  Concretely, we define $\mathcal{L_\text{ela}}$ which penalizes the deviation of the log singular values of $\mathbf{M}_t$ from zero, \emph{i.e.},
\begin{align}
\mathcal{L_\text{ela}}=\big\|\log \mathbf{\Sigma}_t\big\|_\text{F}^2.
\label{eq:loss_ela}
\end{align}

\textbf{Multi-level registration.} 
For  ease of optimization, the point registration process during pose decomposition is empirically achieved in a hierarchical manner. Specifically,  $\text{MLP}_\text{r}$ in Eq.~(\ref{eq:rotate})  is a multi-level pyramid network~\cite{DBLP:journals/corr/abs-2205-12796}. We denote $\mathbf{x}_t^{(k)}$ as the transformed coordinate of the output of the $k$-th  pyramid level at time $t$, where $k=1,\cdots, K$.  Then $\mathbf{x}_t^{(k)}$ is obtained by a hierarchical deformation over the initial coordinate $\mathbf{x}_t^{(0)}$, 
\begin{align}
\mathbf{x}_t^{(k)}=\prod_{j=1}^k s_t^{(j)}\tilde{\mathbf{R}}_t^{(j)}\mathbf{x}_t^{(0)}+\tilde{\mathbf{T}}_t,
\label{eq:klevel}
\end{align}
where $\tilde{\mathbf{T}}_t\equiv\mathbf{T}_t$ denotes the translation. $s_t^{(k)}$ and  $\tilde{\mathbf{R}}_t^{(k)}$ are obtained by the $k$-th level output of $\text{MLP}_\text{r}$.

Let $\tilde{\mathbf{G}}_t^{(k)}=\scalemath{0.7}{
\begin{pmatrix}
\prod_{j=1}^k s_t^{(j)}\tilde{\mathbf{R}}_t^{(j)}\mathbf{x}_t^{(0)}&\tilde{\mathbf{T}}_t\\
0 & 1 
\end{pmatrix} }$, by substituting  $\tilde{\mathbf{G}}_t^n$ in Eq.~(\ref{eq:chamfer})  with $\tilde{\mathbf{G}}_t^{n,(k)}$ which transforms from the point $\mathbf{x}_t^{n,(0)}$, we could formulate the $k$-th chamfer distance loss function, termed $\mathcal{L}_\text{cd}^{(k)}$.

We reformulate $\tilde{\mathbf{R}}_t$ by $\tilde{\mathbf{R}}_t=\prod_{k=1}^K\tilde{\mathbf{R}}_t^{(k)}$, and compute  ${\mathbf{R}}_t$ based on $\tilde{\mathbf{R}}_t$ following Eq. (\ref{eq:singular_value}). In summary, Eq.~(\ref{eq:rotate}) estimates multi-level $\tilde{\mathbf{R}}_t^{(k)}$ instead of directly estimating $\tilde{\mathbf{R}}_t$.

The total loss function $\mathcal{L}$ is summarized as,
\begin{align}
\mathcal{L}= \sum_{k=1}^K\mathcal{L}_\text{cd}^{(k)}+\mathcal{L_\text{ela}}+\mathcal{L}_\text{cyc},
\label{eq:loss_total}
\end{align}
where the  consistency loss $\mathcal{L}_\text{cyc}$ is composed by a 2D part and a 3D part, which will be detailed  in Sec.~\ref{sec:loss}.

\subsection{Discussions for Multi-object Scenarios}
\label{subsec:step2general}
Monocular reconstruction methods usually assume a single target object in the scene~\cite{DBLP:journals/corr/abs-2206-15258} or utilize multiple video sequences that share the same individual~\cite{DBLP:conf/cvpr/YangVNRVJ22}. Yet real scenarios are likely to contain multiple target objects, where  object occlusions and individual differences are both common and challenging. We demonstrate that RPD could adapt to multi-object scenarios.

\textbf{Object occlusions.} We find our framework could handle object occlusions under certain circumstances.  Given multiple  target objects, we design an anti-occlusion silhouette reconstruction loss to deal with the object occlusion issue, as will be detailed in Sec.~\ref{suppsec:implementation}. 

\textbf{Individual differences.} In complex scenarios with multiple non-rigid objects, different individuals of the same category could vary in height,  volume, or scale.  Instead of constructing several independent canonical models for different objects, we allow these individuals to  share the canonical model and  improve the data efficiency. \textbf{1)} This is partially realized by $\Sim(3)$ with the scaling factor $s_t$  (incorporated in Sec.~\ref{subsec:quadratic_pose_field}) which adjusts the size at local regions when performing the point matching in Eq.~(\ref{eq:chamfer}). \textbf{2)} When combined with the linear skinning weights, the ability to handle scale differences of our reconstruction method is further enhanced by shrinking or stretching the control points/bones. Qualitative results  in Sec.~\ref{sec:qualitative} evaluate the effectiveness of handling individual differences in multi-object scenarios.

\section{Experiments}
\label{sec:experiments}

Our experiments are conducted with monocular RGB videos from challenging DAVIS~\cite{DBLP:journals/corr/abs-1905-00737},   OVIS~\cite{DBLP:journals/ijcv/QiGHWLBBYTB22}, AMA~\cite{DBLP:journals/tog/VlasicBMP08}, and Casual~\cite{DBLP:conf/cvpr/YangVNRVJ22} datasets. We  detail  dataset settings   and implementation details in Sec.~\ref{sec:implementation}. We  compare the proposed RPD with  state-of-the-art methods including Nerfies~\cite{DBLP:conf/iccv/ParkSBBGSM21}, ViSER~\cite{DBLP:conf/nips/YangSJVCLR21}, and BANMo~\cite{DBLP:conf/cvpr/YangVNRVJ22} qualitatively in Sec.~\ref{sec:qualitative} and quantitively in Sec.~\ref{sec:quantitive}. In addition, we perform analytical experiments in Sec.~\ref{sec:ablation} to verify the advantage of each component in RPD. More details of network architectures and visualizations are further provided in Sec~\ref{suppsec:implementation}.

\begin{figure*}[t!]
\centering
\includegraphics[width=0.87\textwidth]{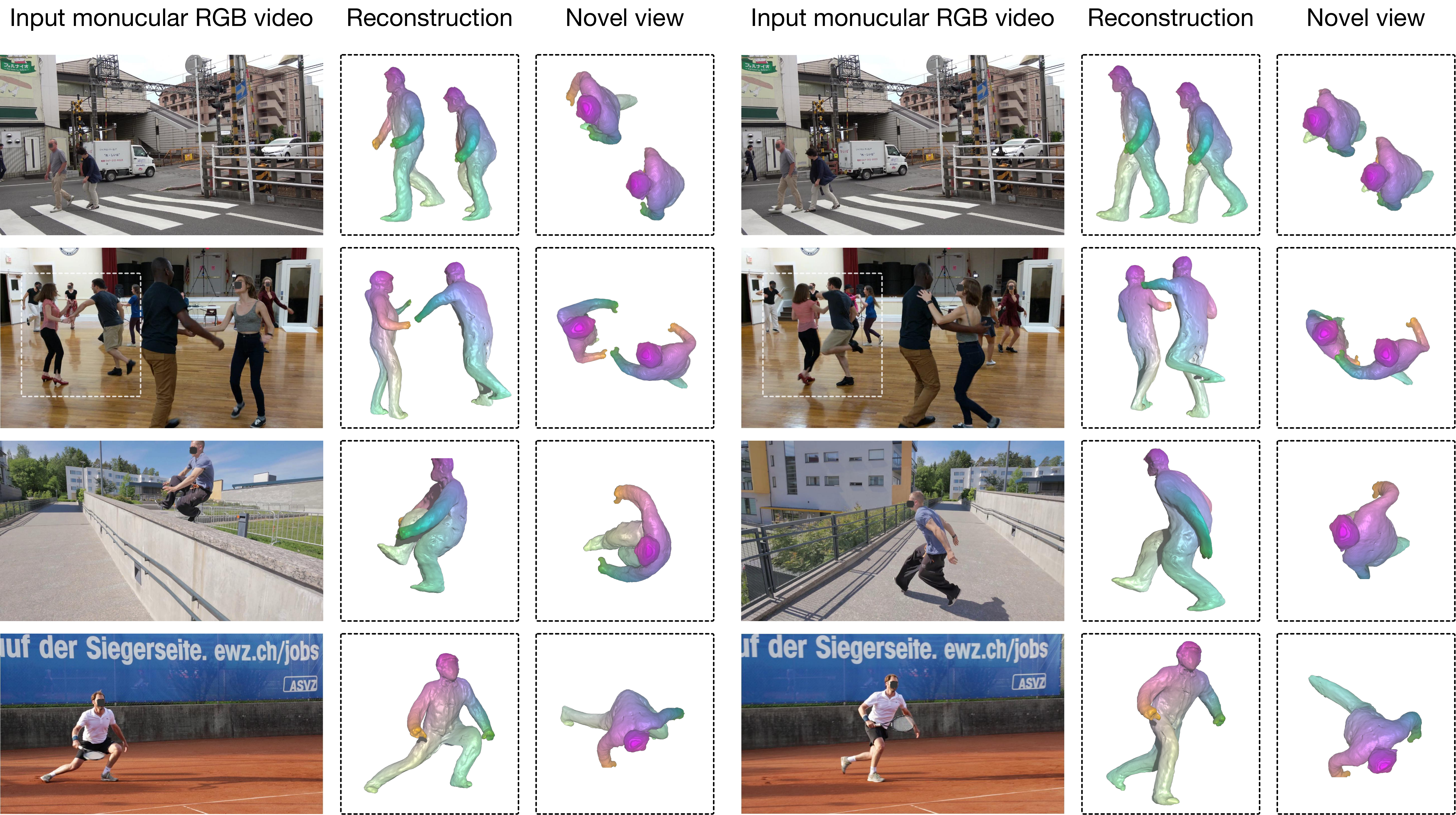}
\caption[]{Reconstruction  of humans trained with $15$ video sequences (without needing SMPL~\cite{DBLP:journals/tog/LoperM0PB15} or pre-trained CSE~\cite{DBLP:conf/nips/NeverovaNSKLV20}/DensePose~\cite{DBLP:conf/cvpr/GulerNK18}). Every two groups in the same row are collected from the same video sequence. Exhibited results cover complicated scenes  containing multiple persons with differences in body size, large pose changes, fast movements like parkour, etc. All these examples share a common canonical space.  A topdown novel view is provided for each group. }
\label{pic:human}
\vskip-0.05in
\end{figure*}

\subsection{Experimental Settings}
\label{sec:implementation}
\textbf{Datasets.} DAVIS~\cite{DBLP:journals/corr/abs-1905-00737} and OVIS~\cite{DBLP:journals/ijcv/QiGHWLBBYTB22} are  video datasets for object segmentation and they both provide dense 2D instance-level object annotations. Concretely, DAVIS  contains 150 videos and $376$ annotated objects.  OVIS covers $25$ categories and collects $901$ video sequences with an average length of $12.77$ seconds. In addition,  OVIS contains more challenging videos that record non-rigid objects with complex motion patterns, rapid deformation, and heavy occlusion. AMA~\cite{DBLP:journals/tog/VlasicBMP08} dataset contains articulated videos with ground-truth meshes and  could be utilized for computing quantitive results. To make a fair comparison with existing methods,   ground-truth object silhouettes are utilized for training, and we adopt the video group named ``$\mathrm{swing}$''. Time synchronization and camera extrinsics are not used for training. The Casual dataset is collected by BANMo~\cite{DBLP:conf/cvpr/YangVNRVJ22} and contains casually captured animals and humans.

\textbf{Implementation details.} We use the similar architecture with $8$ layers for volume rendering as in NeRF~\cite{DBLP:conf/eccv/MildenhallSTBRN20}, but deform observed objects into the canonical space. We initialize $\text{MLP}_\text{SDF}$ as  an approximate unit sphere~\cite{DBLP:conf/nips/YarivKMGABL20}   centered at $\mathbf{o}_t$ (Eq.~(\ref{eq:rotate})).
The dimension of each  latent  code embedding is chosen as $\bm{\omega}_t\in\mathbb{R}^{64}$ in Eq.~(\ref{eq:nerf}), $\mathbf{g}_t\in\mathbb{R}^{128}$ in Sec.~\ref{subsec:quadratic_pose_field}, and $\bm{\varphi}_t\in\mathbb{R}^{128}$ in Eq.~(\ref{eq:rotate}). The time-variant latent code is represented by a Fourier embedding~\cite{DBLP:conf/eccv/MildenhallSTBRN20}. $\text{MLP}_\text{g}$ in Sec.~\ref{subsec:quadratic_pose_field} is a $8$-layer network that takes as input  the Fourier embedded $\mathbf{g}_t$. In addition,
$\text{MLP}_\text{r}$ in Eq.~(\ref{eq:rotate}) is a multi-level network consisting of $9$ pyramid levels~\cite{DBLP:journals/corr/abs-2205-12796} (\emph{i.e.}, $K=9$). Inspired by BANMo~\cite{DBLP:conf/cvpr/YangVNRVJ22}, we reduce the reconstruction  complexity  with linear skinning weights and parameterize $\mathcal{M}_{t\to *}$ and $\mathcal{M}_{*\to t}$ in Eq.~(\ref{eq:warping2}) based on $25$ control points.  We follow ViSER~\cite{DBLP:conf/nips/YangSJVCLR21} that  learns approximate pixel-surface embeddings (which are not category-specific) to obtain reasonable initial poses based on the optical flow.

\begin{figure*}[t!]
\centering
\includegraphics[width=0.905\textwidth]{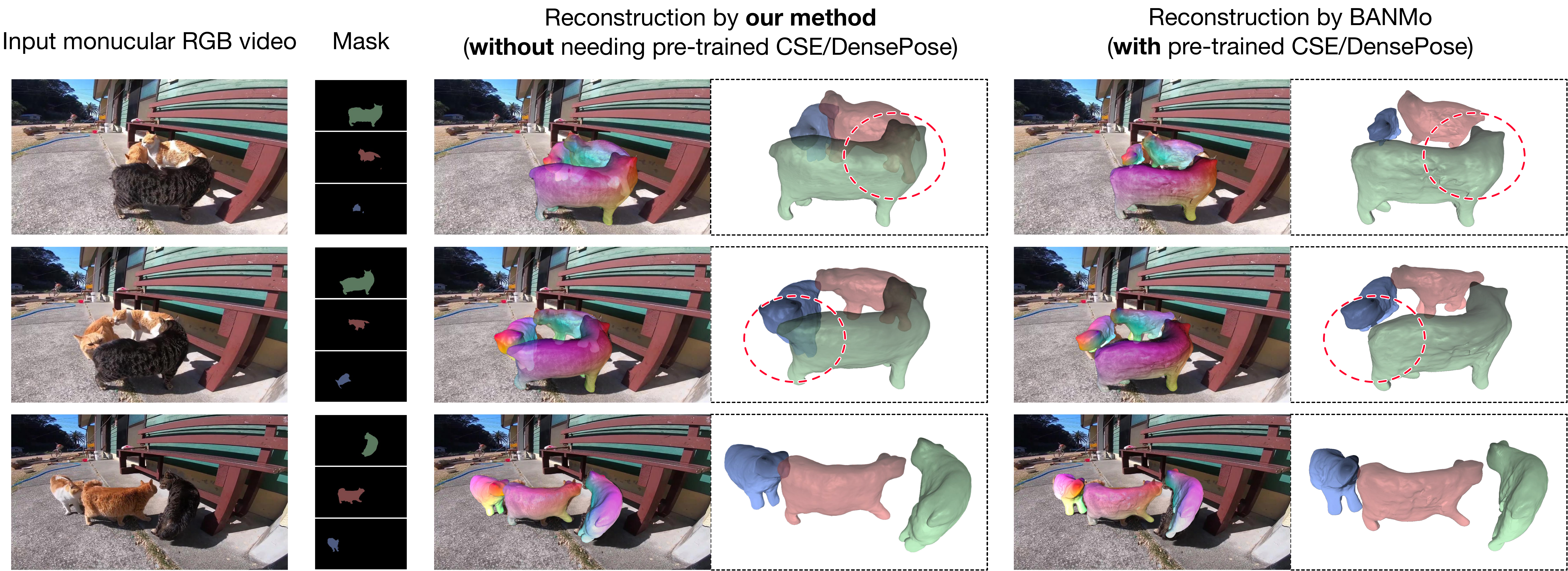}
\caption[]{Qualitative comparison of our method (without needing pre-trained CSE~\cite{DBLP:conf/nips/NeverovaNSKLV20}/DensePose~\cite{DBLP:conf/cvpr/GulerNK18}) vs BANMo~\cite{DBLP:conf/cvpr/YangVNRVJ22} (with pre-trained CSE/DensePose). The experiment reconstructs three highly-occluded cats that walk in a circle. Segmentation masks are also illustrated.  Red dashed circles highlight the regions where RPD outperforms BANMo. Best view in color and zoom in.}
\label{pic:cat}
\end{figure*}

\begin{figure*}[t!]
\centering
\includegraphics[width=0.93\textwidth]{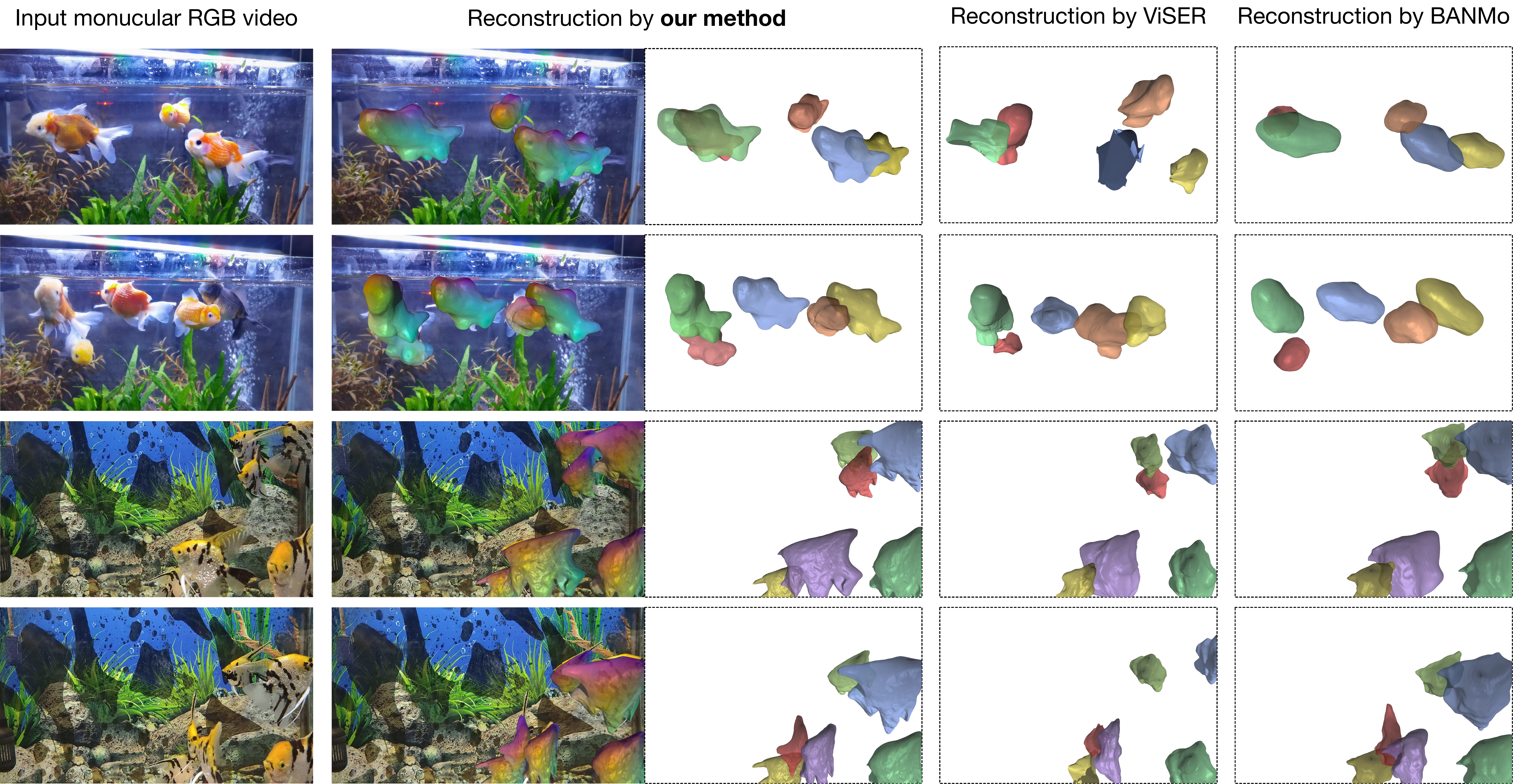}
\caption[]{Qualitative comparison of our method with prior art ViSER~\cite{DBLP:conf/nips/YangSJVCLR21} and BANMo~\cite{DBLP:conf/cvpr/YangVNRVJ22}. We provide two groups of experiments that reconstruct fast-moving fish with complex motion patterns and self-occlusion. Each group is trained with \textbf{only $1$ video sequence} with $11$ seconds, $114$ frames (top $2$ rows) and $12$ seconds, $120$ frames (bottom $2$ rows), respectively. Best view in color and zoom in.}
\vskip-0.08in
\label{pic:fish}
\end{figure*}

\textbf{Optimization details.} For  DAVIS and OVIS datasets, we adopt the provided  segmentation masks when calculating the silhouette reconstruction loss. These  annotated masks are unfilled at occluded pixels, hence the discussed anti-occlusion strategy is still necessary especially given that both datasets contain heavy object occlusion. For the Casual dataset, differently, we predict  segmentation masks by an off-the-shelf network PointRend~\cite{DBLP:conf/cvpr/KirillovWHG20}. For all datasets, we adopt VCN-robust~\cite{DBLP:conf/nips/YangR19} to compute the optical flow as required for optimization in Sec.~\ref{subsec:optimization}. During training, we adopt OneCycle~\cite{smith2019super} as the learning rate scheduler with the initial value, maximum value, and final value being $2\times10^{-5}$, $5\times10^{-4}$, and $1\times10^{-4}$, respectively. For each experiment, we observe that the model could achieve high-fidelity performance by training  20k iterations, which takes approximately $9$ hours on a single V$100$ GPU. To kick-start with a reasonable initial pose, we follow ViSER~\cite{DBLP:conf/nips/YangSJVCLR21} that adopts optical flow for learning approximate pixel-surface embeddings (which are not category-specific). But due to our proposed designs, our performance significantly outperforms 
ViSER as will be compared in Sec.~\ref{sec:qualitative}.

\subsection{Qualitative Results}
\label{sec:qualitative}

To evaluate the effectiveness of our method, we choose hard video samples from the datasets to cover complex object deformations/movements, large pose changes, multi-object scenes with object occlusions, and multiple individuals of the same category yet with  scale differences.

\textbf{Qualitative results of humans.} 
We first consider the non-rigid reconstruction of humans without using off-the-shelf templets (\emph{e.g.}, SMPL~\cite{DBLP:journals/tog/LoperM0PB15}).  Fig.~\ref{pic:human} provides our illustration results of the human reconstruction, trained with  $5$ video sequences from the DAVIS dataset, and $10$ video sequences from the Casual dataset. All reconstructed individuals share a joint canonical space. Different from Nerfies~\cite{DBLP:conf/iccv/ParkSBBGSM21} that assumes the root  pose of an object is compensated by background-SfM, or BANMo~\cite{DBLP:conf/cvpr/YangVNRVJ22} that relies on pre-trained human-specific pose priors (\emph{e.g.}, DensePose),  our RPD optimizes the per time object pose. From the first two rows of demonstrated results, we observe that RPD adapts to individual differences such as different heights, fat/skinny, and local shape scales.  The reconstruction illustration in the following two rows indicates that the pose estimation by registration is relatively accurate even in scenes with large movements such as parkour and playing tennis.

\textbf{Qualitative results of occluded cats.} 
We then illustrate the reconstruction of  cats in Fig.~\ref{pic:cat}, which is jointly trained by $1$ video sequence from the OVIS dataset and $3$ video sequences from the Casual dataset. In addition, we compare our RPD with the state-of-the-art approach BANMo~\cite{DBLP:conf/cvpr/YangVNRVJ22}. Differently, BANMo relies on pre-trained category-specific pose embeddings for the pose estimation, however, ours does not leverage such priors. We observe from the illustration that our reconstructed meshes are realistic and preserve geometric details,  also showing fewer wrinkles on cat bodies. Besides, our superiority in articulated reconstruction is even more highlighted when considering object occlusions. As shown in the illustration, three cats are heavily occluded when walking around, while segmentation masks usually do not capture the occluded regions. Under the circumstance, BANMo fails to predict or reason the portions behind other objects, while our method presents accurate and realistic reconstruction in occluded regions.

\textbf{Qualitative results of single-video fish.} 
To demonstrate the advantage of our RPD on generic object categories such as invertebrates and fish, where off-the-shelf pre-trained pose networks are unavailable. In Fig.~\ref{pic:fish}, we exhibit the reconstruction result of several fish, and each experiment is trained with only $1$ video sequence ($11$ or $12$ seconds) from the OVIS dataset. The reconstruction is performed in a  multi-object scenario with more challenging dynamics, \emph{e.g.}, multiple objects  with heavy occlusion and complex motion patterns.   We compare our reconstruction results with prior art including BANMo~\cite{DBLP:conf/cvpr/YangVNRVJ22}  and ViSER~\cite{DBLP:conf/nips/YangSJVCLR21} by downloading their open-sourced codebases. Similar to our method, both baseline methods utilize 2D supervision of videos including segmentation masks and optical flows. We find that removing the pre-trained CSE embeddings in BANMo simply leads to catastrophic failures. As a result, for BANMo, we still preserve its original implementation of dense pose priors. For a fair comparison, we perform mesh upsampling and smoothing for ViSER to improve its resolution during rendering. It is observed that BANMo might degenerate into a smooth shape with scarce geometric details, which is probably due to the lack of pose priors of the proper object category. Our method clearly outperforms both baseline methods by the holistic articulated shape reconstruction and the local preservation of details, showing the ability to reconstruct generic object categories even  under limited data.

\subsection{Quantitive Results}
\label{sec:quantitive}

\begin{table}[t]
	\centering
	\tablestyle{1pt}{1}
	\resizebox{0.98\linewidth}{!}{
		\begin{tabular}{l|c|ccc|ccc}
			\shline
			 \multirow{2}*{Method} &\multirow{2}*{Training Data}  &\multicolumn{3}{c|}{ AMA-$\mathrm{swing}$ data} &\multicolumn{3}{c}{ $\mathrm{Hands}$ data } \\&&
			\makecell{CSE Pose}
			& \makecell{CD ($\downarrow$)}
			& \makecell{F@$2\%$ ($\uparrow$)}&\makecell{Real Pose}
			& \makecell{CD ($\downarrow$)}
			& \makecell{F@$2\%$ ($\uparrow$)}\\
			\hline
			Nerfies~\cite{DBLP:conf/iccv/ParkSBBGSM21}  &Single-video& \cmark&22.6 & 13.2&-&-&-  \\
			BANMo~\cite{DBLP:conf/cvpr/YangVNRVJ22} & Single-video & \cmark&{9.4} & {56.8} & \cmark & 10.3 & 39.5\\
			\rowcolor{gray!13}
			RPD &Single-video & \cmark &\textbf{9.0}  & \textbf{57.4}&-&-&- \\
			\hline
			BANMo & Single-video & \xmark&28.3 & 10.2 & \xmark &25.1 &14.4   \\
			\rowcolor{gray!13}
			RPD &Single-video & \xmark & \textbf{11.4} & \textbf{52.7} &\xmark & \textbf{13.0} & \textbf{30.9}\\
			\hline
			ViSER~\cite{DBLP:conf/nips/YangSJVCLR21} &Multi-video& \cmark&{15.7} & {52.2}&\cmark&16.8&21.3 \\
			BANMo &Multi-video& \cmark&9.1 & 57.0&\cmark &7.5&49.6\\
\rowcolor{gray!13}
			RPD &Multi-video & \cmark & \textbf{8.5}  & \textbf{58.7} &-&-&-   \\
			\hline
			ViSER  &Multi-video& \xmark&{27.7} & {10.3} &\xmark&27.3&10.1\\
			BANMo &Multi-video& \xmark&24.2 & 12.5 &\xmark&18.7&19.4\\
\rowcolor{gray!13}
			RPD &Multi-video& \xmark &{\textbf{10.1}}   & \textbf{53.9}  &\xmark&\textbf{8.2}   &\textbf{47.0} \\
			\shline
		\end{tabular}
	}
\vskip0.04in
\caption{Quantitive results on single/multi-video AMA-$\mathrm{swing}$~\cite{DBLP:journals/tog/VlasicBMP08}  and $\mathrm{Hands}$~\cite{DBLP:conf/cvpr/YangVNRVJ22} datasets. Evaluation metrics:  Chamfer distance (CD) and F-score (F@$2\%$) averaged over all frames.}
\vskip-0.03in
\label{table:ama}
\end{table}

\begin{table}[t]
	\centering
	\tablestyle{7.5pt}{1.08}
	\resizebox{0.98\linewidth}{!}{
		\begin{tabular}{l|c|c|c|c}
			\shline
			\hskip0.02in Method & Training Data &CSE Pose
			& {Dataset}
			& \makecell{mIoU (\%, $\uparrow$)}\\
			\hline
			BANMo~\cite{DBLP:conf/cvpr/YangVNRVJ22} & Multi-video & \cmark&DAVIS (Human)  &72.7  \\
			BANMo & Multi-video & \xmark&DAVIS (Human)  &35.1  \\
			\rowcolor{gray!13}
			RPD & Multi-video &\xmark&DAVIS (Human)  &\textbf{75.9}   \\
			\hline
			BANMo & Multi-video & \cmark & OVIS (Cats)  &{76.2}  \\
			BANMo & Multi-video & \xmark & OVIS (Cats)  &{53.6}  \\
			\rowcolor{gray!13}
			RPD  & Multi-video &\xmark&OVIS (Cats) &\textbf{87.7}   \\\hline
			BANMo & Single-video &\xmark& OVIS (Fish) &{49.1}  \\
			ViSER~\cite{DBLP:conf/nips/YangSJVCLR21} & Single-video &\xmark& OVIS (Fish) &{40.4}  \\
			\rowcolor{gray!13}
			RPD & Single-video &\xmark&OVIS (Fish) &\textbf{70.2}   \\
			\shline
		\end{tabular}
	}\vskip0.04in
\caption{Quantitative results of mIoU (\%), computed with  estimated 2D foreground masks and  ground truth masks.}
\label{table:miou}
\vskip-0.12in
\end{table}

We consider chamfer distance and F-score as  3D metrics to quantify the performance on  AMA~\cite{DBLP:journals/tog/VlasicBMP08}  and $\mathrm{Hands}$~\cite{DBLP:conf/cvpr/YangVNRVJ22}. Chamfer distance calculates the average distance between ground-truth and points at the reconstructed surface by finding the nearest neighbour matches. We  report the F-score at distance thresholds $2\%$ of the longest edge of the axis-aligned object bounding box~\cite{DBLP:conf/cvpr/TatarchenkoRRLK19}. Experimental results are provided in Table~\ref{table:ama}. By comparison, our RPD achieves better quantitive performance in all settings, and shows significant improvement when the pre-trained category-specific CSE poses~\cite{DBLP:conf/nips/NeverovaNSKLV20}  and the ground truth poses are not available, \emph{e.g.},  $53.9$ vs $12.5$ (AMA-$\mathrm{swing}$ data) and $47.0$ vs $19.4$ ($\mathrm{Hands}$ data) for the multi-video setting.  Besides, based on the results of our RPD with/without CSE, we find that RPD also benefits from a proper pose prior for initialization.

Besides  3D  metrics, we also report 2D mIoU (averaged over all frames) between the annotated segmentation masks and estimated masks. To avoid mistakenly computing mIoU at occlusion pixels, we only consider two classes (foreground and background) for all datasets. Although mIoU is not a standard  metric for 3D scenes, it  reflects the reconstruction fidelity to some extent, especially for multi-object scenarios where different individuals share the canonical space. Results in Table~\ref{table:miou} indicate that our RPD demonstrates better performance in all three dataset settings. In particular, RPD outperforms both baselines by a large margin when CSE pose initialization for the given object category (\emph{e.g.}, fish)  is not available, such as $70.2$ vs $49.1$ when compared with BANMo. These results show the superiority of our method in capturing high-fidelity geometric details while not relying on category-specific dense pose priors.

\begin{table}[t]
	\centering
	\tablestyle{3pt}{1}
	\vskip-0.13in
	\resizebox{0.98\linewidth}{!}{
		\begin{tabular}{l|ccc|ccc}
			\shline
			 \multirow{2}*{Method}  &\multicolumn{3}{c|}{ AMA-$\mathrm{swing}$ data} &\multicolumn{3}{c}{ $\mathrm{Hands}$ data} \\&
			\makecell{CSE pose}
			& \makecell{CD ($\downarrow$)}
			& \makecell{F@$2\%$ ($\uparrow$)}&\makecell{Real pose}
			& \makecell{CD ($\downarrow$)}
			& \makecell{F@$2\%$ ($\uparrow$)}\\
			\hline
						ViSER~\cite{DBLP:conf/nips/YangSJVCLR21} & \cmark&{15.7} & {52.2}&\cmark&16.8&21.3 \\
			BANMo~\cite{DBLP:conf/cvpr/YangVNRVJ22}  & \cmark&$\;\,$9.1 & 57.0&\cmark &$\;\,$7.5&49.6\\
			\rowcolor{gray!9}
			ViSER + RPD  & \cmark&\textbf{12.2} \tiny{\textcolor{Highlight}{(-3.5)}} & \textbf{53.6} \tiny{\textcolor{Highlight}{(+1.4)}} &-&-&- \\
			\rowcolor{gray!9}
			BANMo + RPD & \cmark & $\;\,$\textbf{8.8} \tiny{\textcolor{Highlight}{(-0.3)}}  & \textbf{58.2} \tiny{\textcolor{Highlight}{(+1.2)}} &-&-&-   \\
			\hline
			ViSER & \xmark&{27.7} & {10.3} &\xmark&27.3&10.1\\
			BANMo & \xmark&24.2 & 12.5 &\xmark&18.7&19.4\\
			\rowcolor{gray!9}
			ViSER + RPD & \xmark & \textbf{13.9} \textbf{\tiny{\textcolor{Highlight}{\underline{(-13.8)}}}}  & \textbf{51.5} \textbf{\tiny{\textcolor{Highlight}{\underline{(+41.2)}}}} &\xmark&\textbf{19.0}  \textbf{\tiny{\textcolor{Highlight}{\underline{(-8.3)}}}} &\textbf{17.3}  \textbf{\tiny{\textcolor{Highlight}{\underline{(+7.2)}}}}  \\
			\rowcolor{gray!9}
			BANMo + RPD & \xmark &{\textbf{11.0}} \textbf{\tiny{\textcolor{Highlight}{\underline{(-14.2)}}}}  & \textbf{53.2} \textbf{\tiny{\textcolor{Highlight}{\underline{(+40.7)}}}} &\xmark&$\;\,$\textbf{8.8}  \textbf{\tiny{\textcolor{Highlight}{\underline{(-9.9)}}}} &$\,\,$\textbf{46.1}  \textbf{\tiny{\textcolor{Highlight}{\underline{(+26.7)}}}} \\
			\shline
		\end{tabular}
	}
\vskip0.04in
\caption{Quantitive results of  applying RPD to ViSER/BANMo, on multi-video AMA-$\mathrm{swing}$~\cite{DBLP:journals/tog/VlasicBMP08}  and $\mathrm{Hands}$~\cite{DBLP:conf/cvpr/YangVNRVJ22} datasets. Evaluation metrics follow Table~\ref{table:ama}.}
\vskip-0.02in
\label{table:rpd_aba}
\end{table}

\subsection{Ablation Studies}
\label{sec:ablation}

\begin{table}[t]
	\centering
	\tablestyle{2pt}{1.07}
	\resizebox{0.98\linewidth}{!}{
		\begin{tabular}{c|cc||c|cc}
			\shline
			{Pose Error}
			& \makecell{Chamfer\\Distance (cm, $\downarrow$)}
			& \makecell{F-score\\@$2\%$ (\%, $\uparrow$)}&{Pose Error}
			& \makecell{Chamfer\\Distance (cm, $\downarrow$)}
			& \makecell{F-score\\@$2\%$ (\%, $\uparrow$)}\\
			\hline
			$0^\circ$ & {9.6}  & {55.9} & $45^\circ$& 9.6 & 55.7\\
			$90^\circ$ & {10.3}  & {54.7} & $135^\circ$ & 16.4 & 50.2 \\
			\shline
		\end{tabular}
	}\vskip0.04in
\caption{Sensitivity to inaccurate initial poses on AMA-$\mathrm{swing}$. Evaluation metrics:  Chamfer distance (cm) and F-score (\%).}
\vskip-0.06in
\label{table:pose_sensi}
\end{table}

\textbf{Appling RPD separately to ViSER/BANMo:} We further demonstrate that  RPD can also serve as a standalone component to rectify root poses for existing methods, \emph{e.g.}, ViSER~\cite{DBLP:conf/nips/YangSJVCLR21} and  BANMo~\cite{DBLP:conf/cvpr/YangVNRVJ22}.
    Results are provided in Table~\ref{table:rpd_aba}. We observe that  combining RPD with baseline methods achieves huge gains when there are \textbf{no}  CSE/real poses. For example, BANMo+RPD obtains 40.7$/$26.7 F-score improvements over BANMo itself when CSE  is not available.

\begin{figure}[t!]
\centering
\includegraphics[width=0.47\textwidth]{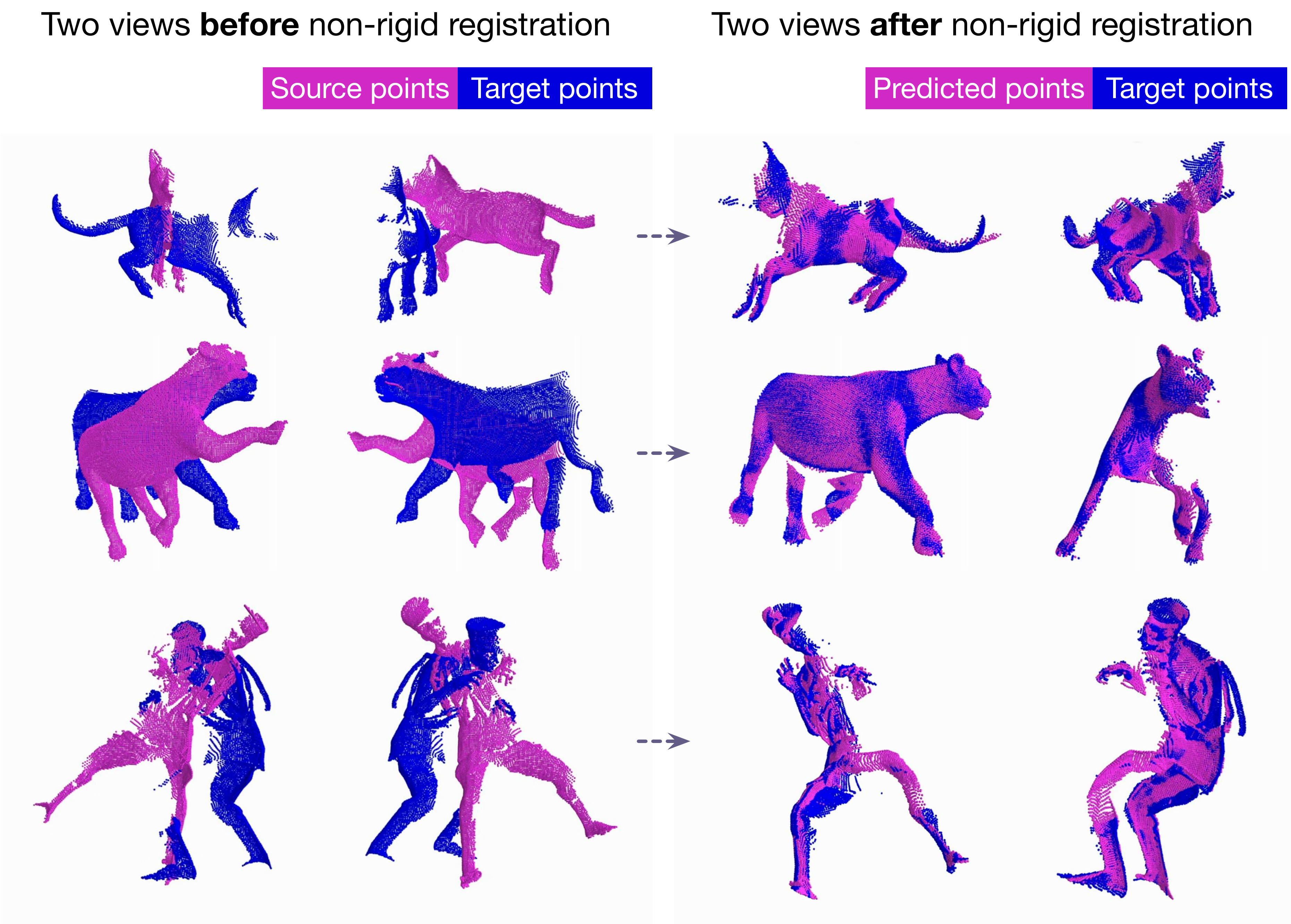}
\caption[]{Evaluation of the multi-level  pyramid structure for non-rigid  point registration, performed on  4DMatch-F~\cite{DBLP:conf/cvpr/LiH22}.   }
\label{pic:regis}
\vskip-0.12in
\end{figure}

\textbf{Sensitivity to  pose initialization.} In Sec.~\ref{sec:implementation}, we  describe the scheme of root pose  initialization. In Table~\ref{table:pose_sensi}, we inject different levels of Gaussian noise into initial poses, resulting in averaged rotation errors of $45^\circ$, $90^\circ$, or $135^\circ$. RPD appears to be relatively stable up to $90^\circ$ pose error.

\textbf{Non-rigid point registration.} We show the effectiveness of multi-level pyramid structure  and perform non-rigid point registration on  4DMatch-F~\cite{DBLP:conf/cvpr/LiH22}, as shown in  Fig.~\ref{pic:regis}. Even  the initial pose error is large (about $90^\circ$), the registration process still achieves a proper matching.

\begin{figure}[t!]
\centering
\vskip-0.13in
\includegraphics[width=0.47\textwidth]{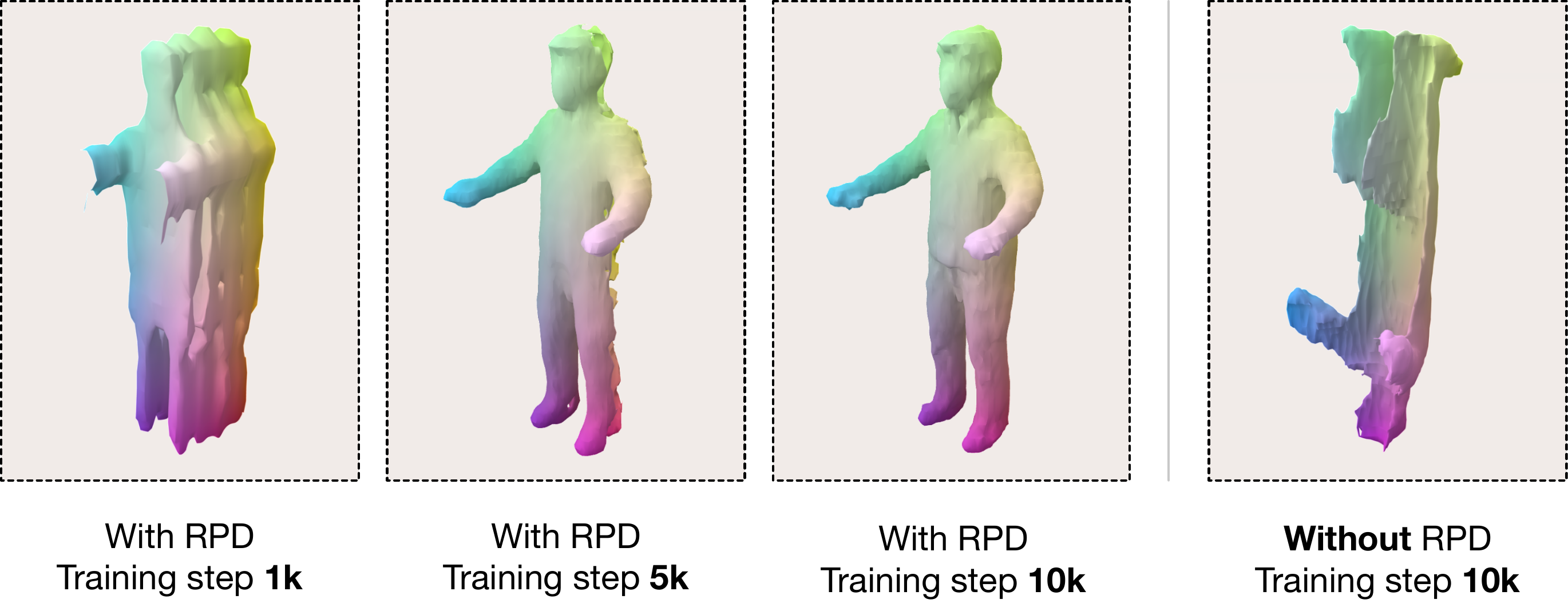}
\caption[]{Meshes of the canonical space with/without RPD at different training steps. }
\label{pic:rpd}
\vskip-0.03in
\end{figure}

\textbf{Reconstruction w.r.t. training steps.} Fig.~\ref{pic:rpd} illustrates meshes that are extracted from the canonical space during training, and  compares the performance with/without RPD. The reconstruction with RPD converges quickly within only 5k training steps, while the reconstruction without RPD fails to handle the front-back human pose ambiguity.

\begin{figure}[t!]
\centering
\includegraphics[width=0.35\textwidth]{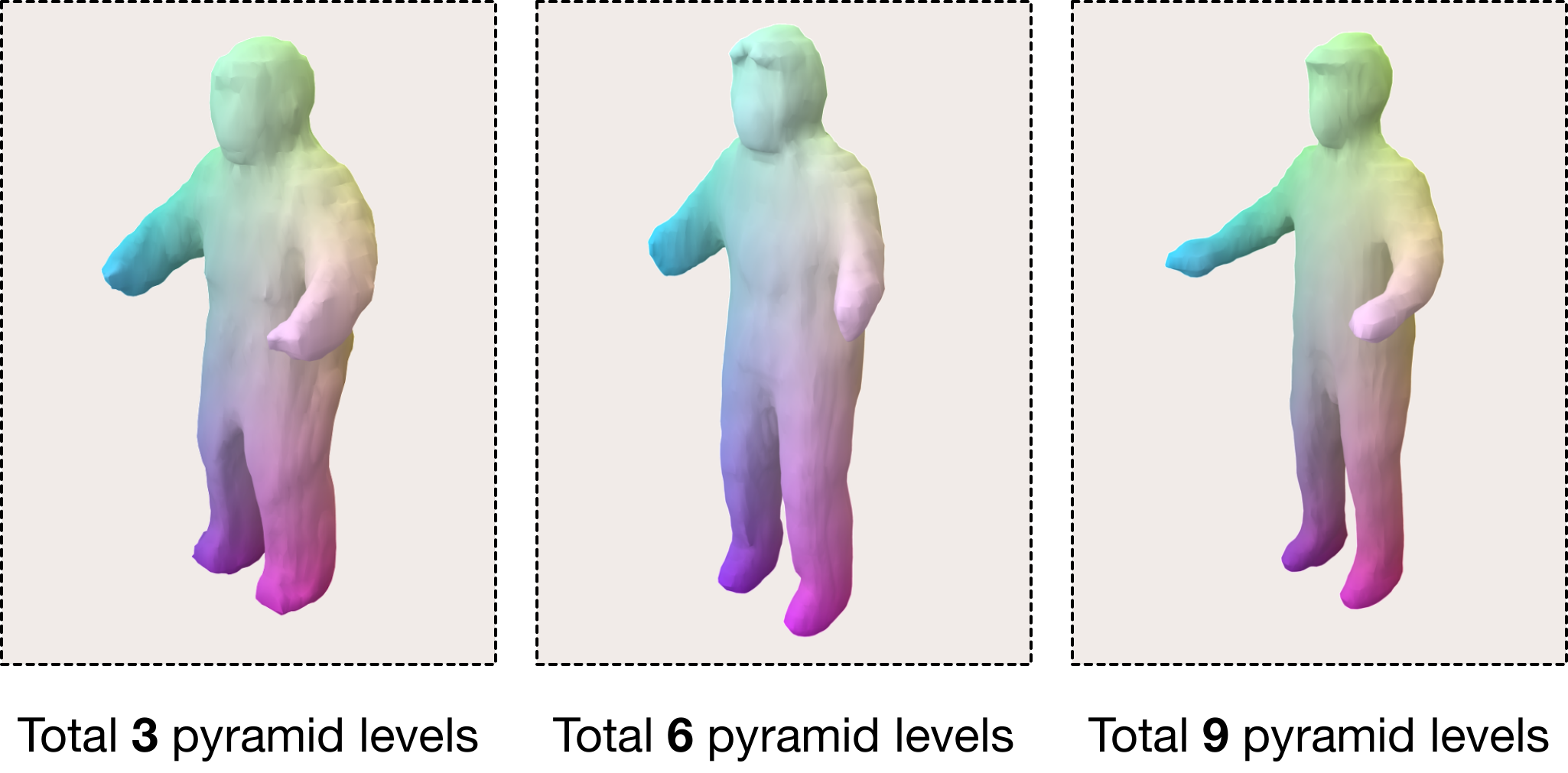}
\caption[]{Meshes of the canonical space under  different pyramid  level settings. All meshes are  extracted at the same training step. }
\label{pic:level}
\vskip-0.03in
\end{figure}

\textbf{Reconstruction w.r.t. pyramid levels.} Fig.~\ref{pic:level}  compares the performance under various pyramid levels. It demonstrates that a larger capability of  pose estimation network (with the increase of  levels) leads to a more elaborate 3D  reconstruction, which evaluates the effectiveness of our RPD.

\begin{figure}[t!]
\centering
\includegraphics[width=0.47\textwidth]{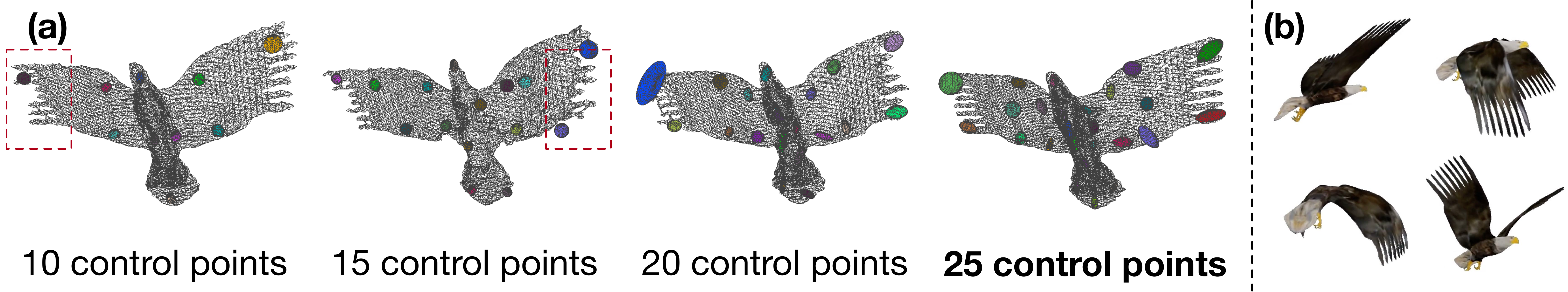}
\caption[]{(a) Different \#control points. (b) Rendered color images.}
\vskip-0.13in
\label{pic:eagle}
\end{figure}

\textbf{Reconstruction w.r.t.  numbers of  control points.} The choice of the number of control points  is  ablated in Fig.~\ref{pic:eagle}(a), where using more control points achieves to a more elaborate  result. We  
 show rendered color images in Fig.~\ref{pic:eagle}(b).

\section{Conclusion}
We propose Root Pose Decomposition (RPD) to reconstruct generic non-rigid objects from monocular videos. RPD decomposes per-frame root pose by learning a dense neural field with local transformations, which are rectified through the point registration to a shared canonical space. RPD has few dependencies on  category-specific templets, background-SfM, or pre-trained dense poses. Meanwhile, RPD demonstrates promising performance for objects with complex deformations/movements, and in multi-object scenarios containing occlusions and individual differences. {Limitation:} The method is limited when only a narrow range of poses is covered throughout input videos. 

\section*{Acknowledgement}
\small{
This work is funded by the Major Project of the New Generation of Artificial Intelligence (No.~2018AAA0102900) and the Sino-German  Collaborative Research Project Crossmodal Learning (NSFC  62061136001/DFG TRR169). Y. Wang and Y. Dong are supported by the Shuimu Tsinghua Scholar Program.}

\clearpage 
\section*{\Large Appendix}
\appendix

\section{More Details and Discussions}
\label{suppsec:implementation}

In this part, we provide  additional implementation details and discussions for  our method and experiment.

\begin{figure}[h]
\centering
\includegraphics[width=0.46\textwidth]{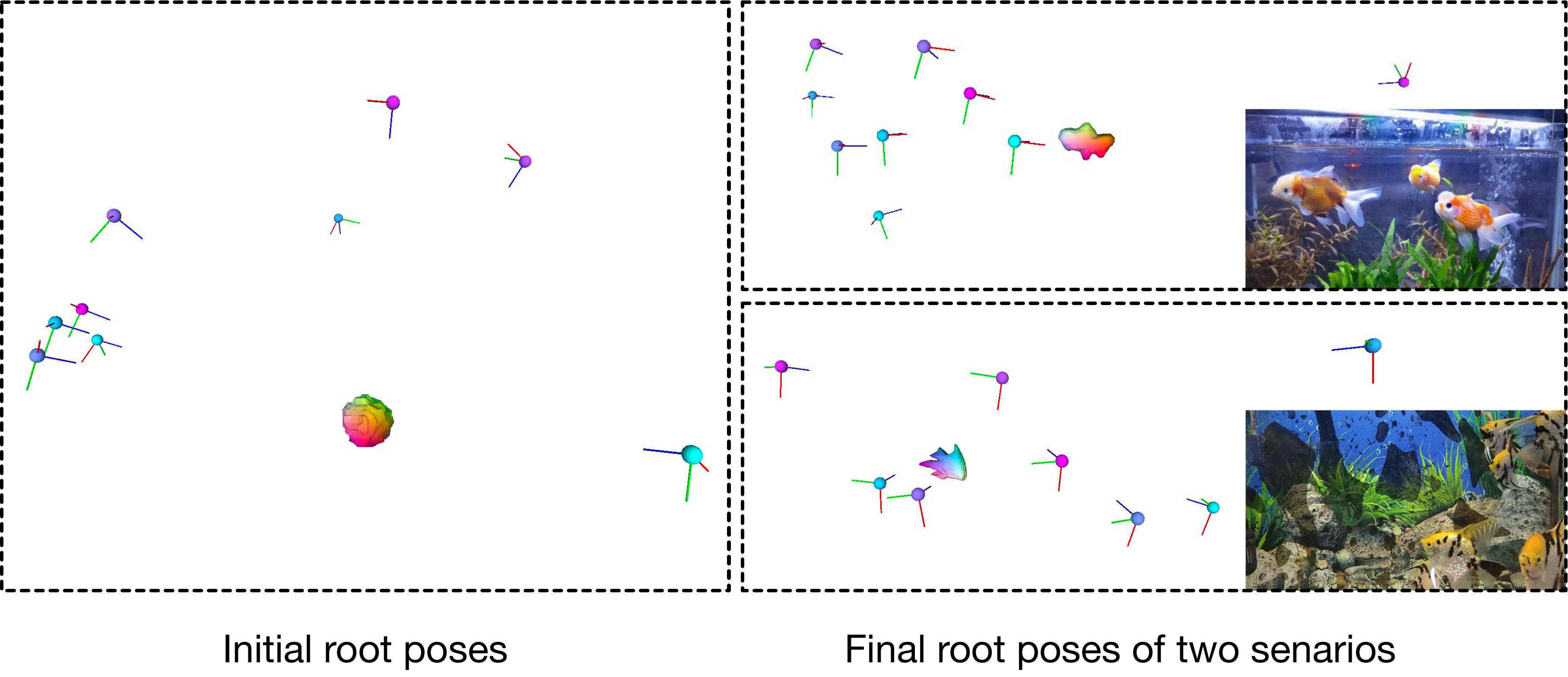}
\caption[]{Illustration of initial root poses  and final root poses with corresponding canonical spaces. Best view in color and zoom in.}
\label{pic:poses}
\end{figure}

\textbf{Root poses during training.} We  provide Fig.~\ref{pic:poses} to compare initial and final root poses, taking two fish scenarios as examples. It is observed that the final root poses could reflect  motion patterns  given  input video sequences.  To kick-start with a reasonable initial pose, we follow ViSER~\cite{DBLP:conf/nips/YangSJVCLR21} that adopts optical flow for learning approximate pixel-surface embeddings (which are not category-specific). But due to our proposed designs, our performance significantly outperforms 
ViSER as compared in Fig.~\ref{pic:fish}.

\begin{figure}[t]
\centering
\includegraphics[width=0.47\textwidth]{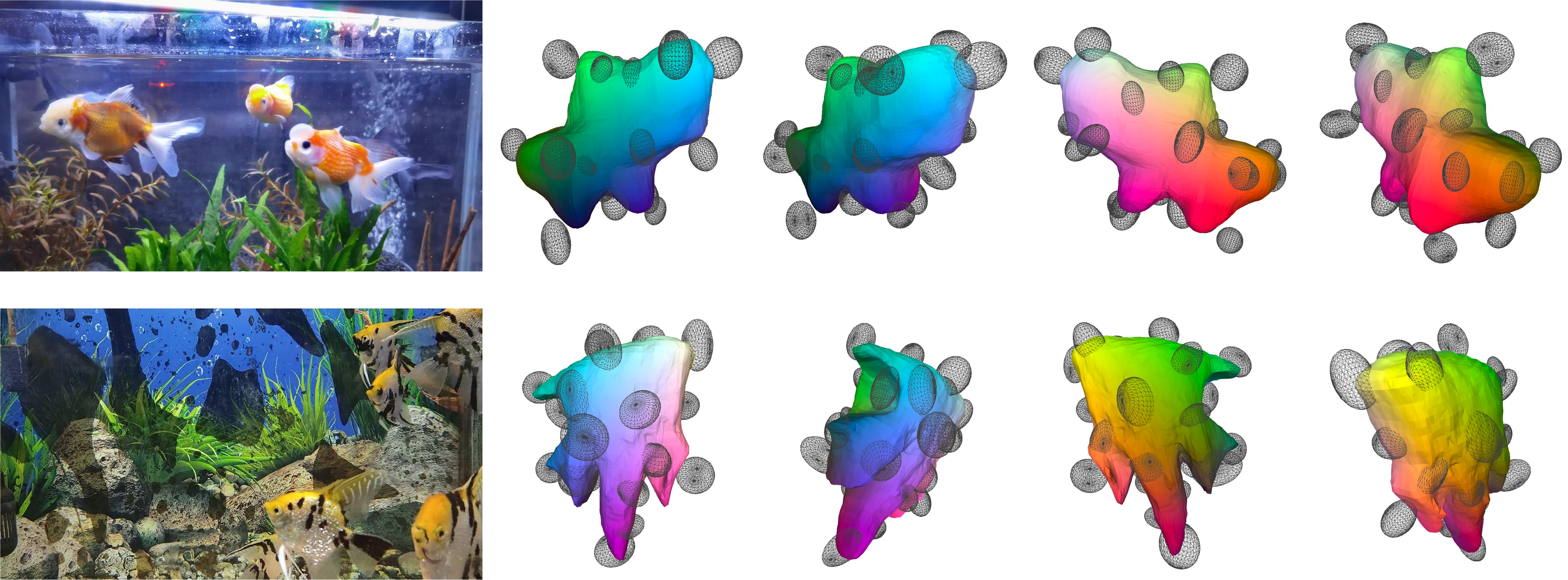}
\caption[]{Illustration of  models in the canonical space on the OVIS dataset with learned control points. }
\label{pic:fish-canonical}
\vskip-0.05in
\end{figure}

\textbf{Visualization of  canonical space.} As mentioned in Sec.~\ref{sec:implementation}, we  adopt $25$ control points when optimizing linear skinning weights. Fig.~\ref{pic:fish-canonical} provides two examples of the learned  models in the canonical space on  OVIS (fish), with also control points. By illustration, we observe that these  canonical models well capture  the geometric shapes of  target objects. Besides, mostly, we find that using $25$/$30$ control points gets similar results. As a result, we follow BANMo~\cite{DBLP:conf/cvpr/YangVNRVJ22}'s default setting ($25$ control points) for fair comparison.

\textbf{Point sampling for registration.} As mentioned in Sec.~\ref{subsec:optimization}, for registration, we sample points in the camera space based on their $\tau^n$ values. Specifically, we keep a per-frame buffer that contains a maximum of $10^4$ points. Old points are removed from the buffer if the maximum volume is exceeded. During training, we discard a ray if its all $\tau^n$ values are lower than $10^{-3}$. We then sample points by the probability  of  a $\mathrm{Softmax}$ output over $\tau^n$ values, with a temperature of $0.01$. 

\textbf{Decomposition into $\Sim(3)$ or $\SO(3)$.} In Sec.~\ref{subsec:quadratic_pose_field}, we leverage dense $\Sim(3)$ with scaling factors $s_t$ to deal with the scale change between different shapes. If we substitute the dense field with $\SO(3)$ by disregarding $s_t$, we find an accelerated convergence process when learning the canonical space, but the root pose might be inaccurate when encountering individual differences, such as the height difference.

\textbf{Why using both decomposed poses and the root pose.}
The deformation field introduces ambiguities that make optimization more challenging, especially when learning skinning weights. We address this issue by maintaining the global transformation, as described in the main paper, \emph{e.g.}, the caption of Fig.~\ref{pic:method}.

\textbf{Object occlusions.} Compared with the multi-view  3D reconstruction, the issue of object occlusion is less explored when only given monocular videos. We demonstrate that the framework could handle object occlusions.  
For a point $\mathbf{x}_*$ on the object surface of the canonical space, 
denote $\mathbf{p}_t\in\mathbb{R}^2$ as the projected 2D pixel that corresponds to $\mathbf{x}_*$ given the  transformation ${\mathbf{G}}_t$, attained by,
\begin{align}
\mathbf{p}_t=\Pi_t{\mathbf{G}}_t^{-1}\mathcal{M}_{*\to t}(\mathbf{x}_*),
\label{eq:pixel}
\end{align}
where $\Pi_t$ is the video-specific projection matrix  of a pinhole camera.

\begin{figure}[t!]
\centering
\includegraphics[width=0.47\textwidth]{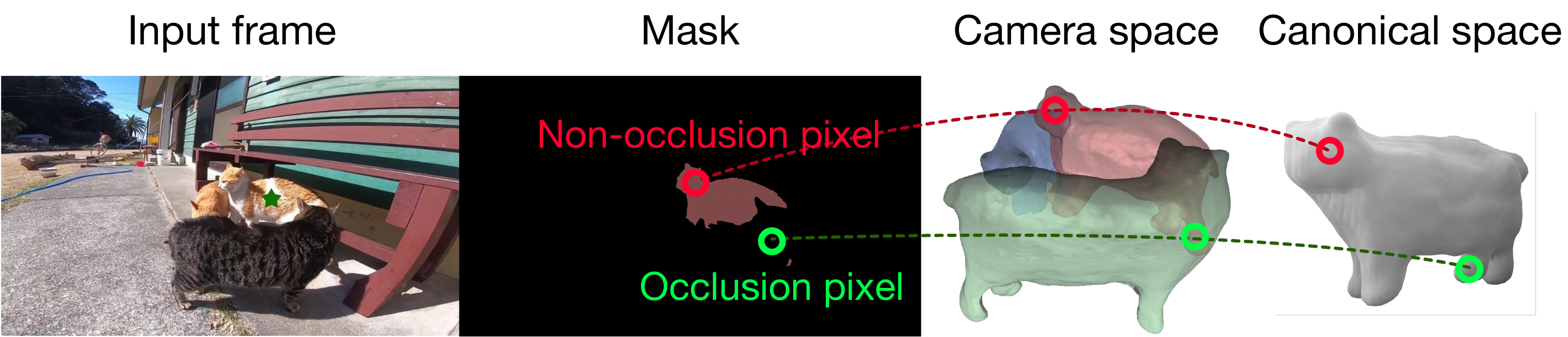}
\caption[]{Illustration of an occlusion pixel that occludes a given target object, in this case an orange and white cat. }
\label{pic:cat-occ}
\vskip-0.12in
\end{figure}

We call $\mathbf{p}_t$ an occlusion pixel if it is outside the annotated 2D mask, as illustrated in Fig.~\ref{pic:cat-occ}. 
In summary, an occlusion pixel is the pixel that occludes a given target object, which comes from a surface point $\mathbf{x}_*$ by Eq.~(\ref{eq:pixel}) yet locates outside the silhouette/mask. Given a target object, denote $\mathbb{U}$ as a set that contains all occlusion pixels.  We design an anti-occlusion silhouette reconstruction loss\footnote{Note that $\tau^n$ is only optimized by Eq.~(\ref{eq:silhouette_loss}) while is temporally frozen when minimizing other terms such as Eq.~(\ref{eq:mini_rotation})  and Eq.~(\ref{eq:cycle_3d_loss}).} defined by,
\begin{align}
\mathcal{L}_\text{sil}=&\sum_{\mathbf{p}_t} \mathbb{A}_{\mathbf{p}_t\notin\mathbb{U}}\Big\|\sum_{n} \tau^n_{\mathbf{p}_t}-\mathbb{I}_{\mathbf{p}_t}\Big\|^2,
\label{eq:silhouette_loss}
\end{align}
where $\sum_{n}\tau^n_{\mathbf{p}_t}$ sums over sampled points along the camera ray that emanates from the pixel $\mathbf{p}_t$, and $\mathbb{I}_{\mathbf{p}_t}\in\{1,0\}$ is an indicator function implying whether $\mathbf{p}_t$ belongs to the segmentation mask of the target object. $\mathbb{A}_{\mathbf{p}_t\notin\mathbb{U}}\in\{1,\alpha\}$ is another indicator function that equals to $1$ if $\mathbf{p}_t\notin\mathbb{U}$ otherwise $\alpha$, where  $\alpha$ is annealing parameter that is initialized to $1$ and decays during the training process.

Similarly, the method is relatively robust to out-of-frame pixels since they are not mistakenly penalized by Eq.~(\ref{eq:silhouette_loss}).

\section{Loss Functions}
\label{sec:loss}

In Sec.~\ref{subsec:optimization}, we basically describe loss functions that are specifically designed/adopted for our method. Here, we detail other loss functions and the formulation of summing all loss functions together. 

Following the standard pipeline~\cite{DBLP:conf/eccv/MildenhallSTBRN20,DBLP:conf/nips/YarivKMGABL20}, we adopt a color reconstruction loss $\mathcal{L}_\text{rgb}$, computed as 
\begin{align}
\mathcal{L}_\text{rgb}=&\sum_{\mathbf{p}_t}\Big\|\sum_{n} \tau^n\mathbf{c}_t\big(\mathcal{W}_{t\to *}(\mathbf{x}_t^n)\big)
-\hat{\mathbf{c}}_t|_{\mathbf{p}_t}\Big\|^2,
\label{eq:rgb_loss}
\end{align}
where $\mathbf{x}_t^n$ is the $n$-th point emanates from the pixel $\mathbf{p}_t$; the color $\mathbf{c}_t(\cdot)$ is defined by Eq.~(\ref{eq:nerf}); and $\hat{\mathbf{c}}_t|_{\mathbf{p}_t}$ denotes the observed color at the pixel $\mathbf{p}_t$.

We further calculate an optical flow loss $\mathcal{L}_\text{flow}$ with a similar formulation with existing methods~\cite{DBLP:journals/corr/abs-2206-15258,DBLP:conf/cvpr/YangVNRVJ22}, 
\begin{align}
\mathcal{L}_\text{flow}=&\sum_{\mathbf{p}_t,(t,t')}\Big\|\mathcal{F}\big(\mathbf{p}_t,t\to t'\big)-\hat{\mathcal{F}}\big(\mathbf{p}_t,t\to t'\big)\Big\|^2,
\label{eq:flow_loss}
\end{align}
where the computed optical flow  $\mathcal{F}(\mathbf{p}_t,t\to t')=\mathbf{p}_{t'}-\mathbf{p}_{t}$, and the observed optical flow $\hat{\mathcal{F}}(\mathbf{p}_t,t\to t')$ is estimated by an off-the-shelf flow network, VCN-robust~\cite{DBLP:conf/nips/YangR19}. Following BANMo~\cite{DBLP:conf/cvpr/YangVNRVJ22}, the pixel $\mathbf{p}_{t'}$ at time $t'$ is obtained by,
\begin{align}
\mathbf{p}_{t'}=\sum_{n}\tau_n\Pi_{t'}\Big(\mathcal{W}_{*\to {t'}}\big(\mathcal{W}_{t\to *}(\mathbf{x}_t^n)\big)\Big),
\label{eq:next_pixel}
\end{align}
where $\Pi_{t'}$ is the video-specific projection matrix (at time $t'$) of a pinhole camera.

For  optimization, we adopt a color reconstruction loss~\cite{DBLP:conf/eccv/MildenhallSTBRN20,DBLP:conf/nips/YarivKMGABL20} and an optical flow loss~\cite{DBLP:conf/cvpr/YangVNRVJ22}. We optimize $\tau^n$ by applying an anti-occlusion silhouette reconstruction loss by Eq.~(\ref{eq:silhouette_loss}). Similar to NSFF~\cite{DBLP:journals/corr/abs-2206-15258}, we maintain the cycle consistency between deformed frames for the monocular reconstruction with a 3D consistency loss given by,
\begin{align}
\mathcal{L}_\text{3D-cyc}=\sum_{i,n} \tau^n \Big\|\mathcal{W}_{*\to t}\big(\mathcal{W}_{t\to *}(\mathbf{x}_t^n)\big)-\mathbf{x}_t^n\Big\|_2^2.
\label{eq:cycle_3d_loss}
\end{align}

The  consistency loss $\mathcal{L}_\text{cyc}$ is composed by the 2D part ($\mathcal{L}_\text{rgb}$, $\mathcal{L}_\text{flow}$, $\mathcal{L}_\text{sil}$) and the 3D part ($\mathcal{L}_\text{3D-cyc}$), namely,
\begin{align}
\mathcal{L}_\text{cyc}=\mathcal{L}_\text{rgb}+\mathcal{L}_\text{flow}+\mathcal{L}_\text{sil}+\mathcal{L}_\text{3D-cyc}.
\label{eq:cycle_loss}
\end{align}

As mentioned in Eq.~(\ref{eq:loss_total}), the total loss function $\mathcal{L}$ is summarized as
\begin{align}
\mathcal{L}= \sum_{k=1}^K\mathcal{L}_\text{cd}^{(k)}+\mathcal{L_\text{ela}}+\mathcal{L}_\text{cyc},
\label{eq:loss_total}
\end{align}
where $\mathcal{L}_\text{cd}^{(k)}$ is the chamfer distance loss function at the $k$-th pyramid level, and $\mathcal{L_\text{ela}}$ denotes the penalty loss for as-rigid-as-possible movement regularization, both of which have been introduced in Eq.~(\ref{eq:chamfer}) and Eq.~(\ref{eq:loss_ela}).

\section{More Visualizations}
\label{sec:limitation}

\begin{figure}[t!]
\centering
\includegraphics[width=0.47\textwidth]{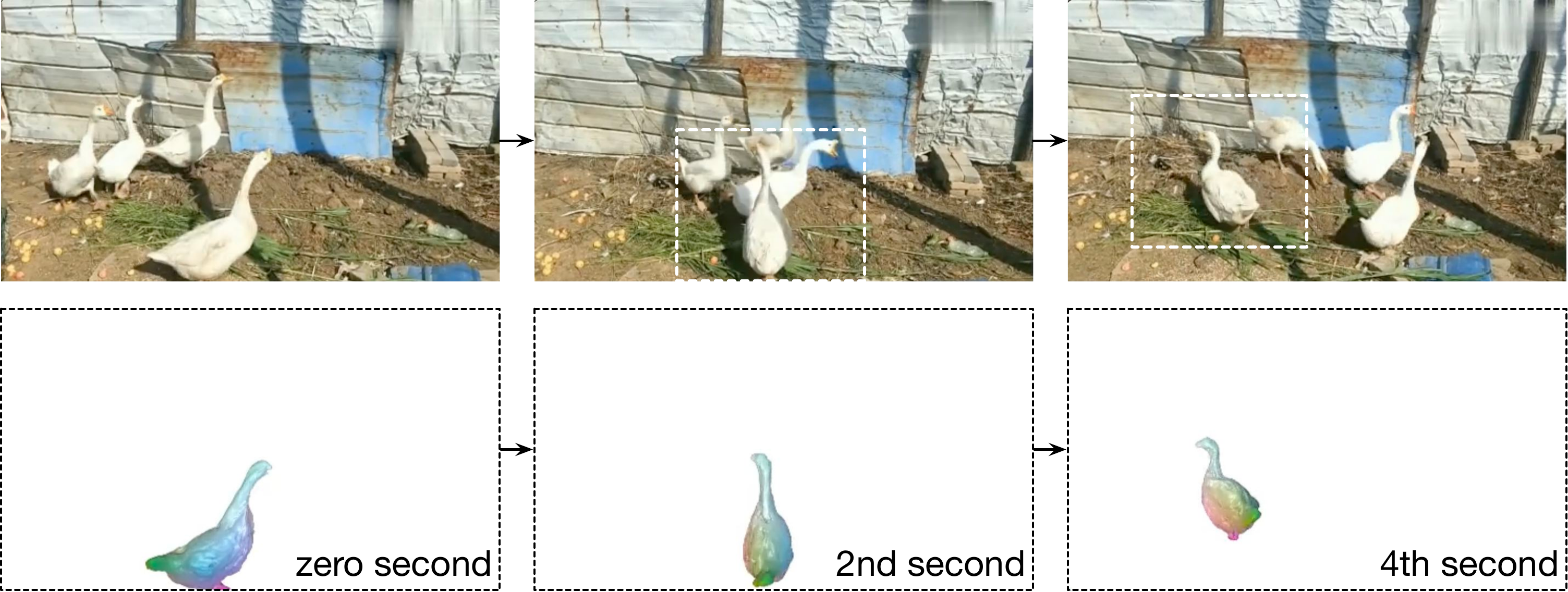}
\caption[]{Illustration of reconstructing a duck. }
\label{pic:duck}
\vskip-0.03in
\end{figure}

We provide Fig.~\ref{pic:duck}  to evaluate RPD on reconstructing ducks.  The experiment is performed by 
jointly using  a 4-second video and a 15-second  complicated video with heavy occlusions.

\begin{figure}[t!]
\centering
\includegraphics[width=0.47\textwidth]{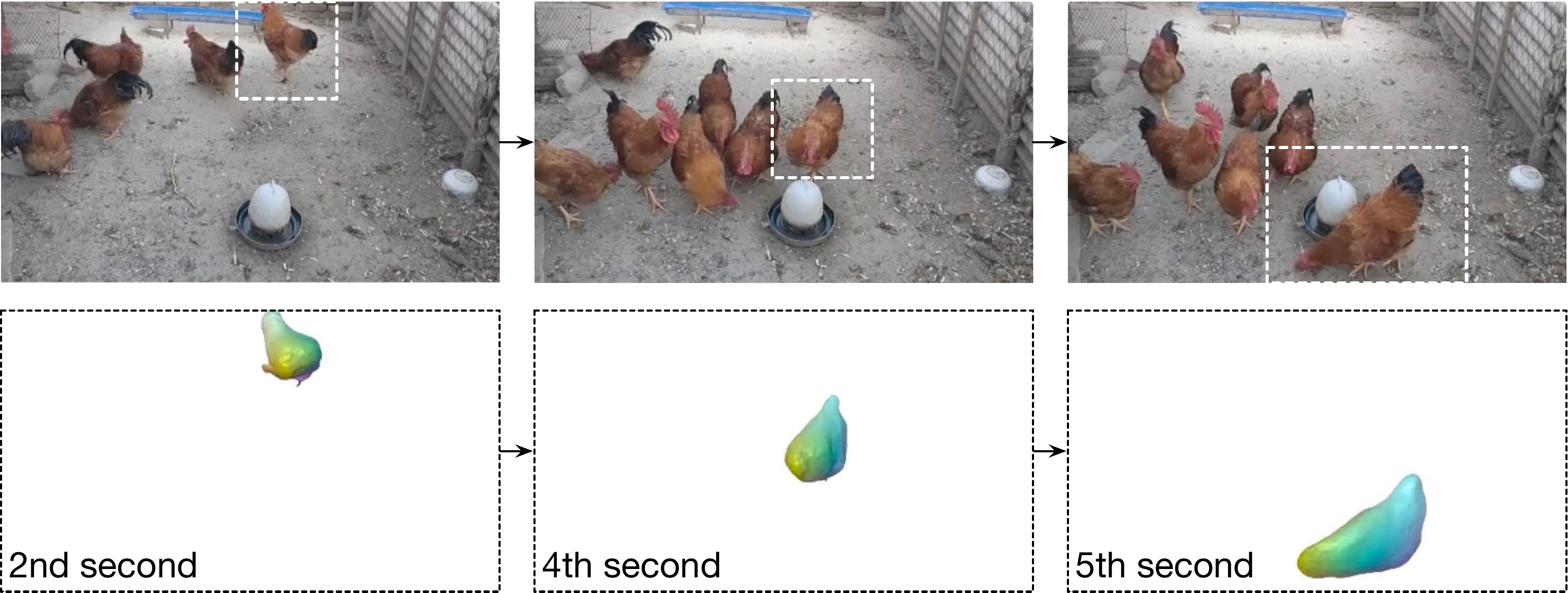}
\caption[]{Illustration of reconstructing a chicken which quickly changes its distance to the camera. }
\label{pic:chicken}
\vskip-0.03in
\end{figure}

As mentioned in Sec.~\ref{subsec:quadratic_pose_field}, for ease of optimization, we  let $\tilde{\mathbf{T}}_t\equiv\mathbf{T}_t$ for all points but learn the per-point rotation matrix $\tilde{\mathbf{R}}_t$. Setting  $\tilde{\mathbf{T}}_t\equiv\mathbf{T}_t$ assumes the camera is at the roughly same distance to the object, which might lead to failure cases when objects quickly running towards  the camera, especially for multi-object cases. To examine the performance, we provided Fig.~\ref{pic:chicken} which reconstructs chickens. We observe that  when a target object rapidly changes its distance to the camera, the reconstruction becomes coarse and the performance is barely acceptable.

\begin{figure}[t!]
\centering
\includegraphics[width=0.4\textwidth]{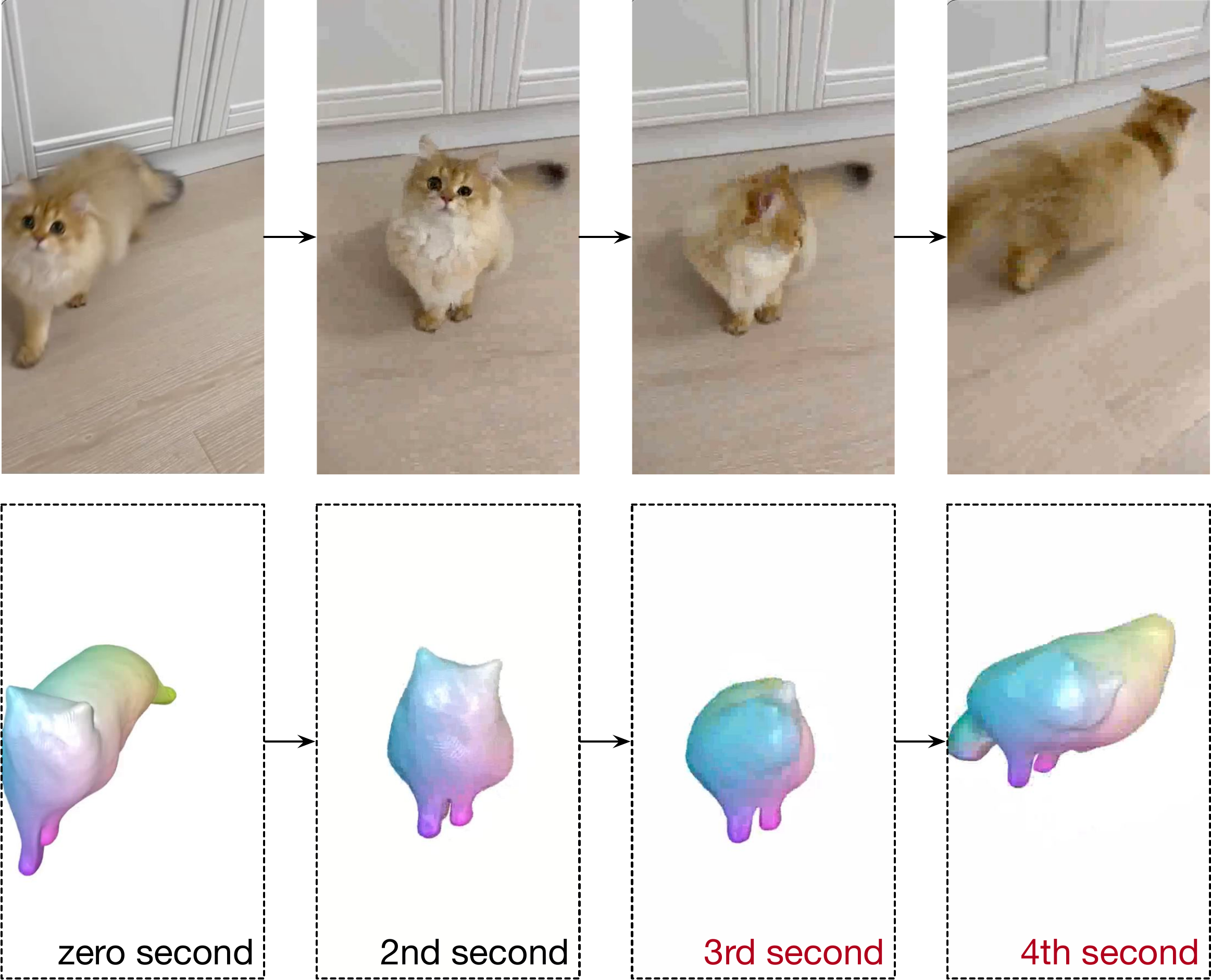}
\caption[]{Failure case: illustration of reconstructing a cat which has a rapid pose change in the 3rd second.  }
\label{pic:cat-fail}
\vskip-0.11in
\end{figure}

A failure case is depicted in Fig.~\ref{pic:cat-fail}, where the root pose encounters a rapid change in the 3rd second, leading to ambiguous pose estimation and confusion in distinguishing the head and tail in the camera space.

{\small
\bibliographystyle{ieee_fullname}
\bibliography{egbib}
}

\end{document}